\documentclass[sigconf,balance=false]{acmart}
\AtBeginDocument{%
  }

\makeatletter
\patchcmd{\@mkbibcitation}{\ref{TotPages}}{10}{}{}
\patchcmd{\@mkbibcitation}{\ref{TotPages}}{10}{}{}
\makeatother

\usepackage{amsthm,amsmath}
\usepackage{mathrsfs}
\usepackage{xspace}
\usepackage{multirow}
\usepackage{enumitem}
\usepackage{makecell}
\usepackage{tikz}
\usepackage{placeins}

\def\name{{\textit{CoAnchor}}\xspace}
\def\base{{\textit{baseST}}\xspace}
\newcommand{\papertitle}{\name: Robust Collaborative Perception under Spatio-Temporal Misalignment via Object-Level Anchors}

\newcolumntype{L}[1]{>{\raggedright\arraybackslash}p{#1}}

\copyrightyear{2026}
\acmYear{2026}
\setcopyright{cc}
\setcctype{by}
\acmConference[MM '26]
  {Proceedings of the 34th ACM International Conference on Multimedia}
  {November 10--14, 2026}
  {Rio de Janeiro, Brazil}
\acmBooktitle{Proceedings of the 34th ACM International Conference on Multimedia
  (MM '26), November 10--14, 2026, Rio de Janeiro, Brazil}
\acmDOI{10.1145/3767308.3835460}
\acmISBN{979-8-4007-2213-4/2026/11}

\begin{document}

\title{\papertitle}

\author{Chi Li}
\orcid{0009-0003-2917-7205}
\affiliation{%
  \department{State Key Laboratory of Networking and Switching Technology}
  \institution{Beijing University of Posts and Telecommunications}
  \city{Beijing}
  \country{China}
}
\email{lichi@bupt.edu.cn}

\author{Rui Lin}
\orcid{0009-0008-0874-0472}
\affiliation{%
  \department{State Key Laboratory of Networking and Switching Technology}
  \institution{Beijing University of Posts and Telecommunications}
  \city{Beijing}
  \country{China}
}
\email{lr\_507@bupt.edu.cn}

\author{Aobo Ji}
\orcid{0009-0008-7811-2594}
\affiliation{%
  \department{State Key Laboratory of Networking and Switching Technology}
  \institution{Beijing University of Posts and Telecommunications}
  \city{Beijing}
  \country{China}
}
\email{bobojassp@bupt.edu.cn}

\author{Dongzhu Xu}
\authornote{Corresponding author.}
\orcid{0000-0003-4053-8772}
\affiliation{%
  \department{State Key Laboratory of Networking and Switching Technology}
  \institution{Beijing University of Posts and Telecommunications}
  \city{Beijing}
  \country{China}
}
\email{xudongzhu@bupt.edu.cn}

\renewcommand{\shortauthors}{Chi Li, Rui Lin, Aobo Ji, and Dongzhu Xu}

\begin{abstract}
Collaborative perception extends the sensing range of a single vehicle by fusing observations from nearby agents, which improves the robustness of autonomous driving.  
In realistic deployments, however, the received collaborator messages are often affected by both communication delay and relative-pose noise, which jointly cause stale observations, spatial misalignment, and unstable feature fusion.
Existing methods usually address these issues from either the spatial or temporal side, but handling them jointly in a unified and efficient manner remains challenging.
In this paper, we propose \name, an anchor-centric spatio-temporal alignment framework for asynchronous collaborative perception. 
Instead of directly reasoning on dense BEV features, \name builds sparse object-level spatio-temporal anchors as a shared interface for pose correction and tightly connects spatial refinement, temporal propagation, and current-time verification within one unified loop, while keeping the overall correction process lightweight. Extensive experiments on both simulated and real-world datasets illustrate that \name remains competitive under clean settings and improves the robustness under joint delay and pose perturbations with a favorable practical accuracy-efficiency trade-off.
\end{abstract}



\maketitle

\section{Introduction} \label{sec_intro}

Multi-agent collaborative perception has become a significant advantage for autonomous driving, because neighboring agents can share complementary observations beyond the field of view of a single vehicle~\cite{coalign,cobevflow,sonata,v2xvit}.
By exchanging rich perception information through the V2X communication technology \cite{cost,where2comm}, the ego vehicle can better detect distant, occluded, or partially visible objects, thereby improving the scene understanding in challenging traffic environments \cite{v2v4real,roco,feoco}.

\begin{figure}
	\centering
	\includegraphics[width=1\linewidth,angle=0]{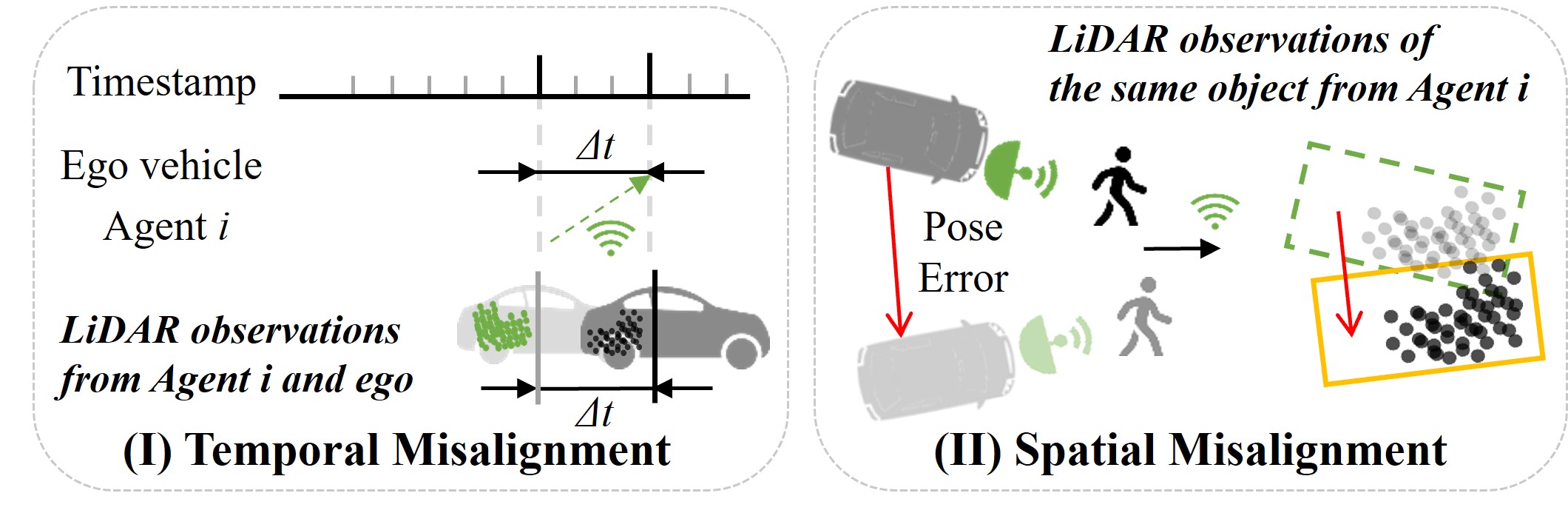}
	\Description{Two panels illustrate cross-agent misalignment. The temporal
	panel shows ego and neighbor LiDAR observations captured at timestamps
	separated by a delay. The spatial panel shows a relative-pose error causing
	the two agents' point clouds and bounding boxes for the same pedestrian to
	be offset.}
	\vspace{-6mm}
	\caption{Temporal and spatial misalignment between ego vehicle and cooperative agents.}
	\label{fig_intro_scenario}	
	\vspace{-6mm}
\end{figure}

Despite the huge potential, collaborative perception is highly sensitive to the  synchronization of cross-agent communications  \cite{traf_align,cost,freeAlign,syncnet,v2xvit}.
As illustrated in Fig.~\ref{fig_intro_scenario}, in {}{realistic deployments}, the received collaborator messages are rarely perfectly aligned with the ego vehicle's current observation due to {}{two types of perception errors}. 
(i) \textit{Temporal misalignment}: due to communication delay, collaborator messages from neighboring agents are captured earlier than the ego timestamp, so the shared observations may already be stale when they arrive at the ego vehicle. 
(ii) \textit{Spatial misalignment}: localization noise and relative-pose error can place the collaborator observations at incorrect locations in the ego frame. 
More seriously, these two factors interweave in practice. 
If the ego vehicle directly fuses the temporally- and spatially-misaligned collaborator information, {}{this} can easily cause shifted detections, duplicated responses, or motion ghosts in the final object perception.

{}{Existing methods usually address these issues by treating temporal and spatial misalignment separately.} %
\textit{{}{On the one hand}}, {}{spatially oriented} methods \cite{coalign,ermvp,roco,cora} refine cross-agent alignment before feature fusion by correcting noisy relative poses or matching object-level observations across agents. They are effective when collaborator observations remain sufficiently synchronized and when reliable correspondences can still be established. 
\textit{On the other hand}, temporally oriented methods \cite{traf_align,syncnet,lrcp,cobevflow} compensate stale collaborator information by propagating or reconstructing delayed features toward the ego vehicle's current time. 
These methods are strong when the spatial reference is already reliable. Nevertheless, in practical collaborative perception, delay and pose noise are coupled rather than isolated: a biased pose estimate may change the starting point of temporal propagation, while {}{propagated collaborator information} should also be re-evaluated at the current ego time before the object feature fusion process. 
Therefore, simply handling spatial correction and temporal compensation in sequence is often insufficient under realistic spatio-temporal misalignment.
This motivates us to seek a unified representation that can couple delayed cross-agent correspondence, temporal propagation, current-time verification, and feature fusion.

In this paper, we propose \textbf{\name}, an anchor-centric collaborative perception framework for spatio-temporally misaligned messages. 
Our key observation is that, although high-dimensional BEV features are powerful for object detection, they do not explicitly expose cross-agent object correspondence or temporal structure under joint delay and pose perturbations. 
By contrast, a low-dimensional object-level representation can provide a shared interface for jointly reasoning about spatial alignment, temporal propagation, and reliability estimation. 
Based on this observation, we represent delayed collaborator objects as \emph{spatio-temporal anchors}, each of which corresponds to a matched cross-agent object hypothesis together with its motion state and reliability cues over time. {These anchors connect delayed-time correspondence and pose refinement, current-time propagation and verification, closed-loop reliability feedback, and anchor-guided feature fusion within a unified pipeline.}

{}
Compared with directly performing pose corrections in dense BEV feature space, \name's anchor-centric design has the following advantages. \emph{(i)} It couples the temporal propagation with current-time posterior verification tightly, so the collaborator perceptive information can be calibrated before the object fusion, to mitigate the effects of communication delay and pose noise. \emph{(ii)} It mainly acts on sparse anchors rather than repeatedly invoking heavy-cost BEV feature processing modules. Thus, it remains efficient in practice, and one feedback round can provide a better accuracy-efficiency trade-off.

Extensive experiments on OPV2V \cite{opv2v} and real-world V2V4Real \cite{v2v4real} demonstrate that \textit{\name remains competitive in clean settings and becomes consistently more robust under coupled pose perturbation and communication delay}. 
In particular, on V2V4Real under a delay of 200\,ms and pose noise of $(0.6\,\mathrm{m}, 0.6^\circ)$, \name achieves 67.04 AP@0.5, outperforming the state-of-the-art spatio-temporal cascade baseline by 6.00 AP@0.5, corresponding to a 9.83\% relative improvement.

We next summarize the contributions.
{}
{\emph{(i)} We analyze why collaborative perception degrades under joint spatio-temporal misalignment caused by communication delay and relative-pose noise. \emph{(ii)} We propose \name, an anchor-centric framework that uses currently verifiable object-level anchors to tightly connect spatial correction, temporal propagation, and collaborative fusion, instead of treating them as independent stages. \emph{(iii)} Extensive evaluations show that \name achieves improved robustness with a favorable accuracy--efficiency trade-off.}

\section{Related Work}

\textbf{Collaborative perception.}
Multiple agents can improve the scene understanding by exchanging sensory information or intermediate representations~\cite{survey,survey1}.
According to the stage at which information is fused, existing methods are commonly divided into early~\cite{cooper}, intermediate, and late fusion~\cite{6232130,8569832}. 
Intermediate-fusion methods offer a better balance between detection accuracy and transmission cost, and have therefore become the dominant paradigm~\cite{where2comm,cobevt,scope,codyntrust}.
However, practical collaborative perception needs to cope with communication delay, localization noise, and imperfect cross-agent synchronization, which make multi-agent fusion significantly more challenging in real deployments \cite{9732063, 10517450}.

\textbf{Spatial calibration under pose noise.}
A line of research focuses on the degradation caused by relative-pose noise~\cite{fpv-rcnn,v2vnetro,cobevglue}. 
CoAlign~\cite{coalign} formulates collaborative calibration through agent-object pose graph optimization and improves robustness without requiring precise relative-pose supervision. RoCo~\cite{roco} further models pose correction as iterative object matching and pose adjustment, emphasizing the importance of reliable cross-agent correspondences under noisy conditions. 
These methods are effective when observations remain approximately synchronized and cross-agent correspondences are reliable. However, under asynchronous collaboration, spatial refinement alone cannot resolve temporal staleness, and its performance may degrade when the matched landmarks are noisy or incorrect.

\textbf{Temporal compensation under communication delay.}
Another line of work studies how to compensate stale collaborator information under communication delay~\cite{ffnet,cobevflow}. 
SyncNet~\cite{syncnet} reconstructs current-time features from historical observations through temporal feature interaction. TraF-Align~\cite{traf_align} introduces trajectory-aware feature alignment, where low-dimensional motion cues guide deformable attention along temporally aligned object trajectories. 
These methods substantially improve robustness to latency when the delayed collaborator message has already been placed in a reliable spatial frame. However, they usually assume the incoming feature as a valid starting point, so residual pose bias can still be propagated forward and appear as duplicated responses or motion ghosts after fusion.

\begin{figure*}
	\centering
	\begin{minipage}{0.6\linewidth}
		\centering
		\begin{minipage}[t]{0.32\linewidth}
			\centering
			\includegraphics[width=0.8\linewidth,angle=0]{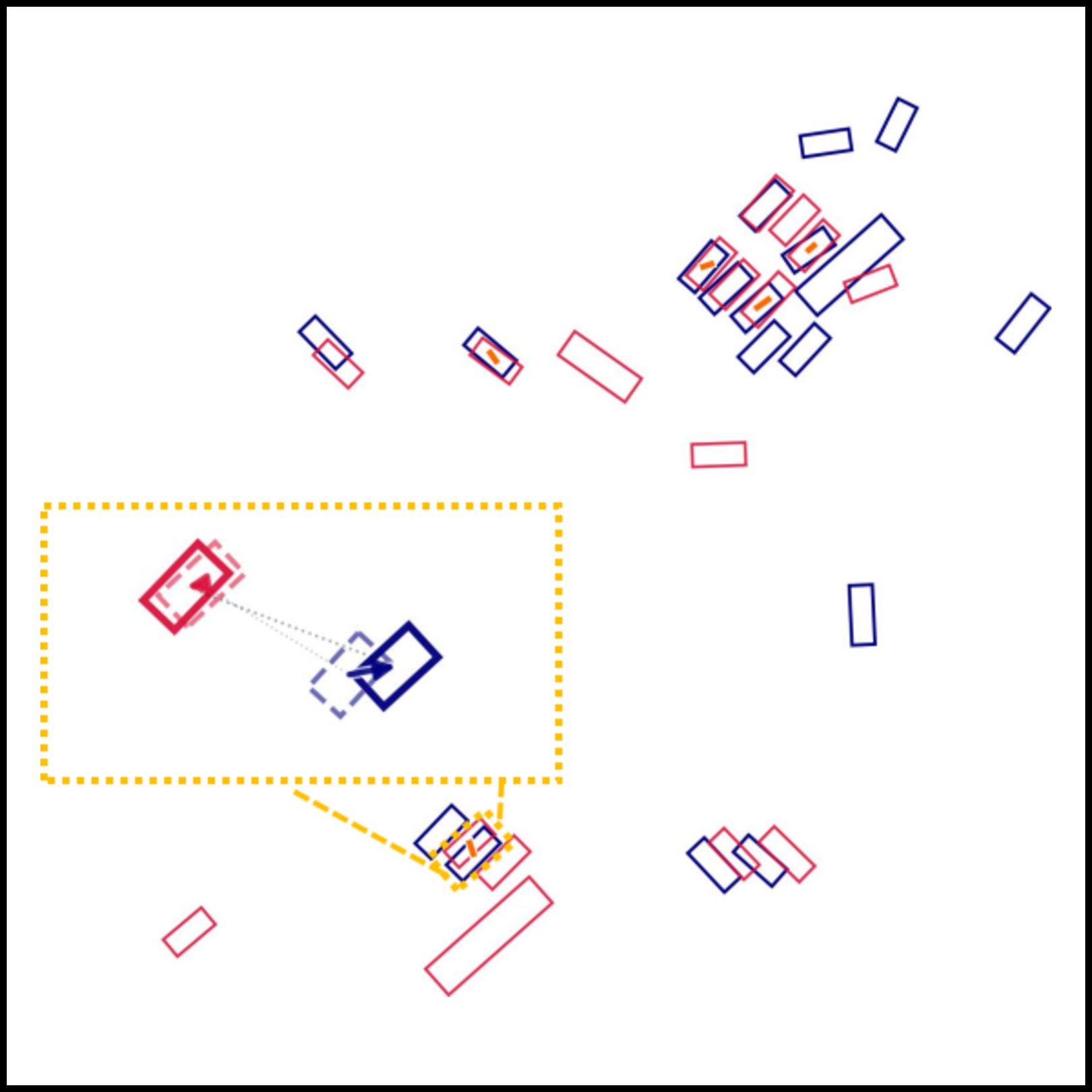}
			\vspace{-2mm}
			\caption*{(a) Raw detection.}
		\end{minipage}
		\begin{minipage}[t]{0.32\linewidth}
			\centering
			\includegraphics[width=0.8\linewidth,angle=0]{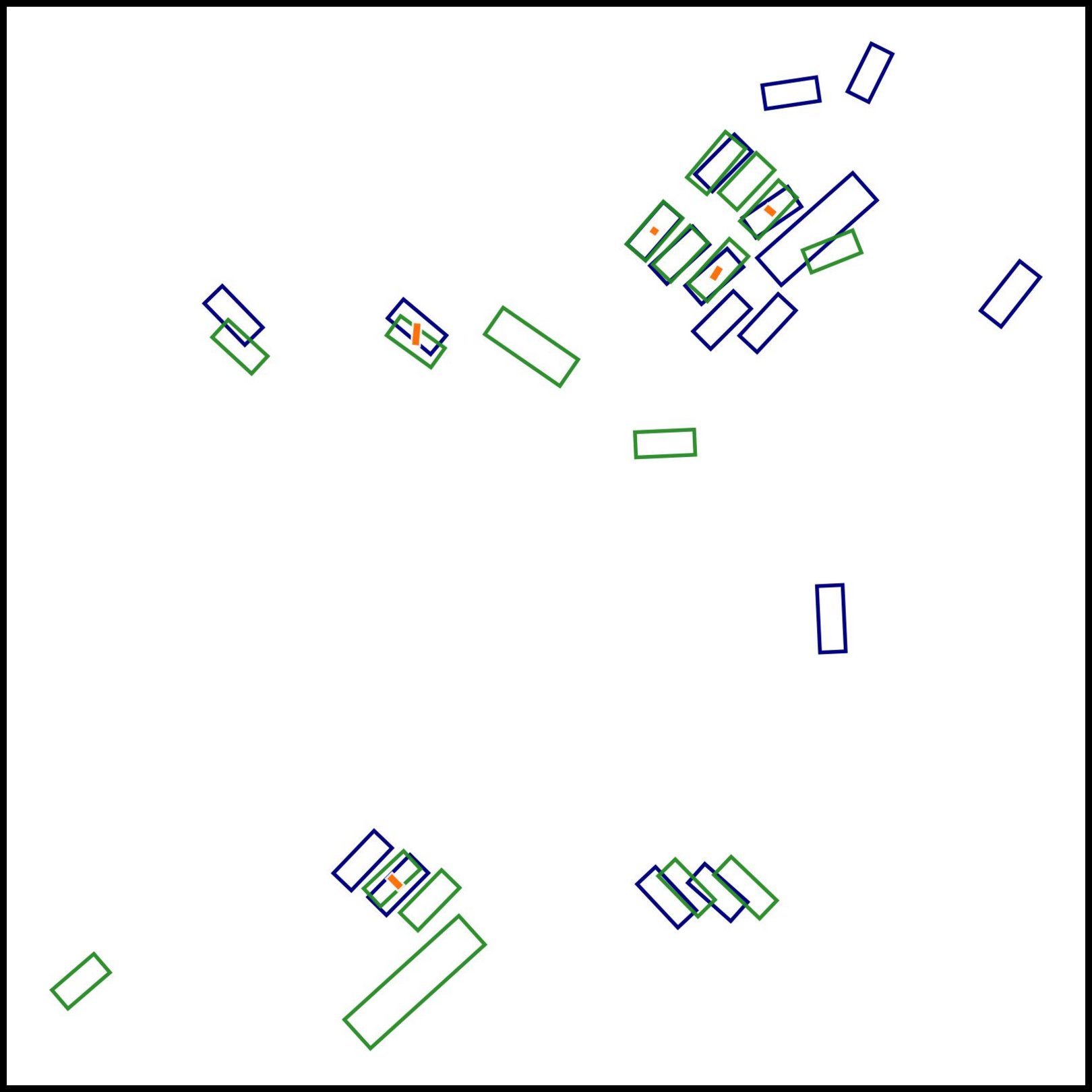}
			\vspace{-2mm}
			\caption*{(b) {}{Compensated result.}}
		\end{minipage}
		\begin{minipage}[t]{0.32\linewidth}
			\centering
			\includegraphics[width=0.8\linewidth,angle=0]{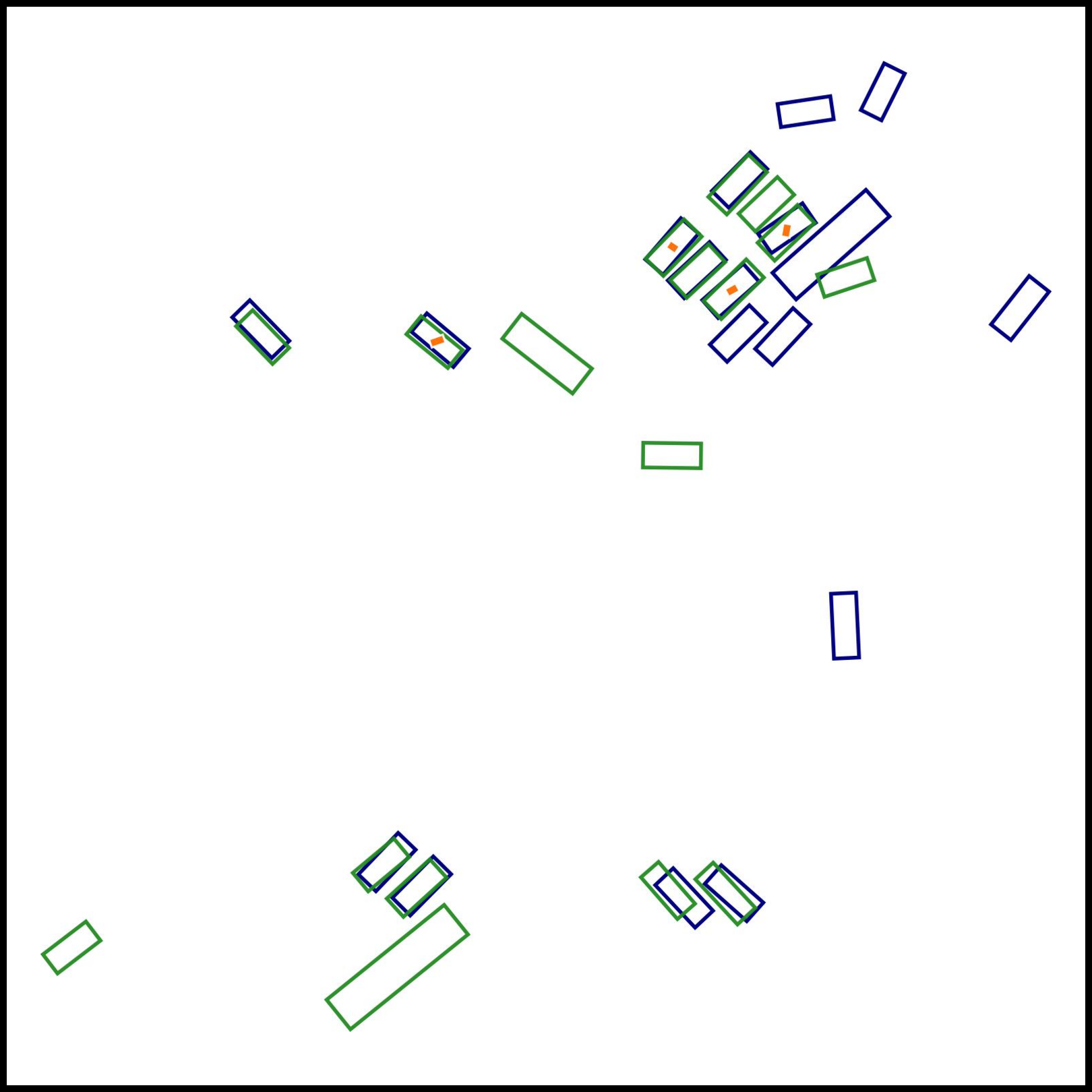}
			\vspace{-2mm}
			\caption*{(c) Ideal detection.}
		\end{minipage}
		{\footnotesize
			\setlength{\tabcolsep}{0pt}
			\renewcommand{\arraystretch}{1.04}
			\newcommand{\legendicon}[1]{\makebox[2.8em][l]{\raisebox{-0.7ex}[0pt][0pt]{#1}}}
			\begin{tabular*}{0.94\linewidth}{@{\extracolsep{\fill}}ll@{}}
				\legendicon{\tikz[baseline=-0.6ex]\draw[red,line width=0.9pt] (0,0) rectangle (0.28,0.18);}Original neighbor box &
				\legendicon{\tikz[baseline=-0.6ex]\draw[red,dashed,line width=0.9pt] (0,0) rectangle (0.28,0.18);}Neighbor box two frames earlier \\
				\legendicon{\tikz[baseline=-0.6ex]\draw[blue,line width=0.9pt] (0,0) rectangle (0.28,0.18);}Current ego box &
				\legendicon{\tikz[baseline=-0.6ex]\draw[blue,dashed,line width=0.9pt] (0,0) rectangle (0.28,0.18);}Ego box two frames earlier \\
				\legendicon{\tikz[baseline=-0.6ex]\draw[green!60!black,line width=0.9pt] (0,0) rectangle (0.28,0.18);}CoAlign-corrected neighbor box &
				\legendicon{\tikz[baseline=-0.6ex]\draw[orange!90!black,line width=1.1pt] (0,0.09) -- (0.32,0.09);}Cluster matching relation \\
				\legendicon{\tikz[baseline=-0.6ex]{
						\draw[-stealth,red,line width=1.0pt] (0,0.05) -- (0.32,0.05);
						\draw[-stealth,blue,line width=1.0pt] (0,0.15) -- (0.32,0.15);
				}}Velocity &
		\end{tabular*}}
		\vspace{-4mm}
			\caption{Showcase of spatial alignment outputs (detected bounding boxes) {}{for raw detections, CoAlign-compensated detections, and ideal detections}.}
			\label{fig_moti_bad_case_coalign}
			\Description{Three bird's-eye-view box plots compare raw detection,
			CoAlign-compensated detection, and ideal alignment. The raw panel
			highlights a mismatched object relation, while the compensated and
			ideal panels show progressively closer overlap between ego and
			neighbor boxes.}
		\end{minipage}
	\begin{minipage}{0.39\linewidth}
		\centering
		\begin{minipage}[t]{0.48\linewidth}
			\centering
			\vspace{0pt}
			\includegraphics[width=0.85\linewidth,angle=0]{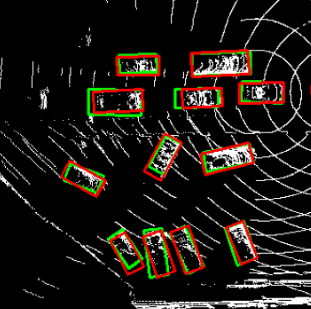}
			\captionsetup{width=0.9\textwidth}
			\caption*{(a) {}{w/o pose noise with a 200\,ms delay.}}
		\end{minipage}
		\begin{minipage}[t]{0.48\linewidth}
			\centering
			\vspace{0pt}
			\includegraphics[width=0.85\linewidth,angle=0]{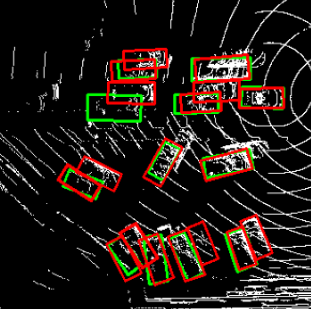}
			\caption*{{}{\shortstack[c]{(b) Delay: 200\,ms\\Pose noise: $\mathbf{(0.6\,\mathrm{m}, 0.6^\circ)}$}}}
		\end{minipage}
		\vspace{-4mm}
		\captionsetup{width=0.9\textwidth}
		\caption{{}{Because of residual spatial errors, many motion ghosts remain even after temporal compensation.}}
		\label{fig_moti_bad_case_traf}
		\Description{Two LiDAR bird's-eye views compare a 200 millisecond delay
		without and with pose noise. Bounding boxes overlap more closely without
		pose noise; with pose noise, several boxes are shifted or duplicated,
		illustrating motion ghosts after temporal compensation.}
		
	\end{minipage}
	\vspace{-4mm}
\end{figure*}

\textbf{Optimization under {}{spatio-temporal} misalignment.}
Recent methods attempt to improve robustness under both spatial and temporal perturbations~\cite{v2xpnp,v2xdgpe,sparsealign}. %
V2X-ViT \cite{v2xvit} exhibits partial robustness to noisy collaboration through transformer-based feature aggregation. CoST  \cite{cost} introduces temporal context by treating historical features as delayed collaborators and retrieving them through deformable sampling. {CoDiff~\cite{codiff} uses conditional diffusion to denoise and progressively refine fused feature representations corrupted by delay and pose errors.} 
Nevertheless, under large spatio-temporal perturbations, these methods still predominantly depend on high-dimensional feature interaction to implicitly reconstruct the aligned scene. 
This implicit strategy becomes unreliable under large spatio-temporal perturbations, because the model has to recover object correspondence, motion changes, and feature misalignment all from noisy high-dimensional features. Consequently, scene recovery may become unreliable when delayed observations are both temporally outdated and spatially biased. 

Different from these approaches, our method is built around \emph{object-level anchors}. Instead of performing all corrections directly in dense BEV feature space, we construct a low-dimensional object-level representation and achieve fusion within one unified pipeline.

\section{Measurement and Motivation} \label{sec_moti}

\subsection{Measurement Setups and Results}

To address the challenge of spatio-temporal asynchrony in collaborative perception, a straightforward strategy is to cascade a spatial alignment module with a temporal compensation module. 
For instance, one can combine CoAlign~\cite{coalign}, which mitigates spatial errors via agent-object pose graph optimization, with TraF-Align~\cite{traf_align}, which compensates communication delay through trajectory-aware feature alignment. 
We denote this cascade solution as \base.

A pilot study shows that such a feed-forward cascade does not provide a stable plug-and-play gain under joint delay and pose noise. 
On V2V4Real, under a fixed 100\,ms delay, inserting CoAlign before TraF-Align reduces AP@0.5 / AP@0.7 from 78.76 / 50.80 to 73.16 / 43.86 even without pose noise. When pose noise of $(0.6\,\mathrm{m}, 0.6^\circ)$ is further introduced under the same delay, the cascaded result drops to 60.68 / 37.39, below the ego-only reference of 62.19 / 40.55.
These results suggest that simply stacking spatial pose refinement and temporal compensation cannot consistently provide trustworthy collaborator evidence for current-time fusion. 
Please refer to Sec.~\ref{sec_eval} {} for more detailed comparisons.

In summary, our study leads to two observations. 
\emph{(i)} CoAlign cannot fully correct the relative pose errors between the ego and the delayed neighbor, because its performance is sensitive to the quality of single-agent detections and the correctness of the matched object pairs. 
\emph{(ii)} TraF-Align can compensate delay once a reasonable spatial reference is available, but it has no explicit mechanism to determine whether the collaborator features remain geometrically trustworthy after pose correction. As a result, residual spatial errors are inherited by the temporal alignment stage and may finally appear as duplicated responses or motion ghosts after propagation.

\subsection{Performance Analysis}

\textbf{Low-quality landmarks and mismatch sensitivity in the spatial alignment module.}
We attribute the instability of CoAlign in realistic scenes to the fact that its pose solver is driven by a limited set of matched detection pairs, whose quality directly depends on the reliability of single-agent detections. 
In practice, the failure does not only come from the lack of sufficient landmarks. 
More significantly, the optimizer can be dominated by \emph{bad} landmarks. We observe two sources of such bad constraints. 
\emph{(i) Low-quality detections}: due to sparse observations, occlusion, and imperfect perception, some detected boxes are themselves inaccurate and therefore provide unreliable geometric references for pose refinement. 
\emph{(ii) Incorrect correspondences}: even when boxes are detected, the matching stage may associate incompatible objects across agents. 
In both cases, the optimization is no longer guided by physically reliable constraints, and a few bad landmarks can bias the global solution and drag the refined pose away from the correct one.

Figure~\ref{fig_moti_bad_case_coalign} presents one representative failure case. In this example, the compensated result is not uniformly poor because the scene itself is unalignable; rather, the main error is caused by one influential mismatched pair, together with other low-quality landmarks that reduce the reliability. %
Once the wrong match is removed, the global alignment becomes visibly better. 
The issue is not that box-based pose refinement is always ineffective, but that its reliability is highly sensitive to the quality of detected landmarks and correctness of matched object pairs.
The same example also reveals a useful cue for diagnosing such bad correspondences. The wrong pair is not only spatially misleading in the delayed frame, but {also }exhibits the worst temporal consistency among all matched pairs in the current frame. %
This suggests that temporal information is useful not only for delay compensation, but {also }for identifying unreliable correspondences that survive single-frame matching. 

\textbf{Residual spatial errors in temporal delay compensation.}
TraF-Align can compensate communication delay effectively only when the incoming neighbor feature has already been placed into a reliable coordinate frame. 
When a biased relative pose is used to warp the delayed neighbor feature into the ego frame, the temporal module starts from a spatially biased input. 
Its temporal propagation itself is still designed to recover delayed information, but it does not explicitly determine whether the incoming representation is already contaminated by residual spatial misalignment. 
As a result, the temporal branch cannot remove such spatial bias on its own: the delay-compensated feature is propagated from a wrong starting point, and the remaining spatial error is further carried into the subsequent feature aggregation stage.

As illustrated in Figure~\ref{fig_moti_bad_case_traf}, these residual spatial errors may finally appear as duplicated responses and motion ghosts after temporal compensation and fusion. 
In other words, the main issue is not that temporal compensation becomes meaningless under pose noise, but that it lacks an explicit mechanism to identify and suppress residual spatial errors. %
This explains why temporal alignment alone may still behave unstably under joint pose noise and communication delay, even though it remains effective in cleaner settings.

The above observations reveal a practical gap in \base. %
The spatial pose-refinement module can improve the relative pose only when the matched detection pairs are reliable, but it cannot guarantee such reliability in realistic scenes. 
The {}delay-compensation module can recover delayed information once the spatial reference is reasonably accurate, but it does not explicitly identify or remove residual spatial errors inherited from the upstream warping stage. 
Therefore, the final object fusion may simultaneously suffer from imperfect temporal recovery and remaining spatial bias. 
These limitations motivate a more tightly coupled design, in which spatial pose refinement, temporal propagation, and current-time reliability verification are connected within a unified closed loop.

\section{Design} \label{sec_design}

	\begin{figure*}
		\centering
		\includegraphics[width=0.9\linewidth,angle=0]{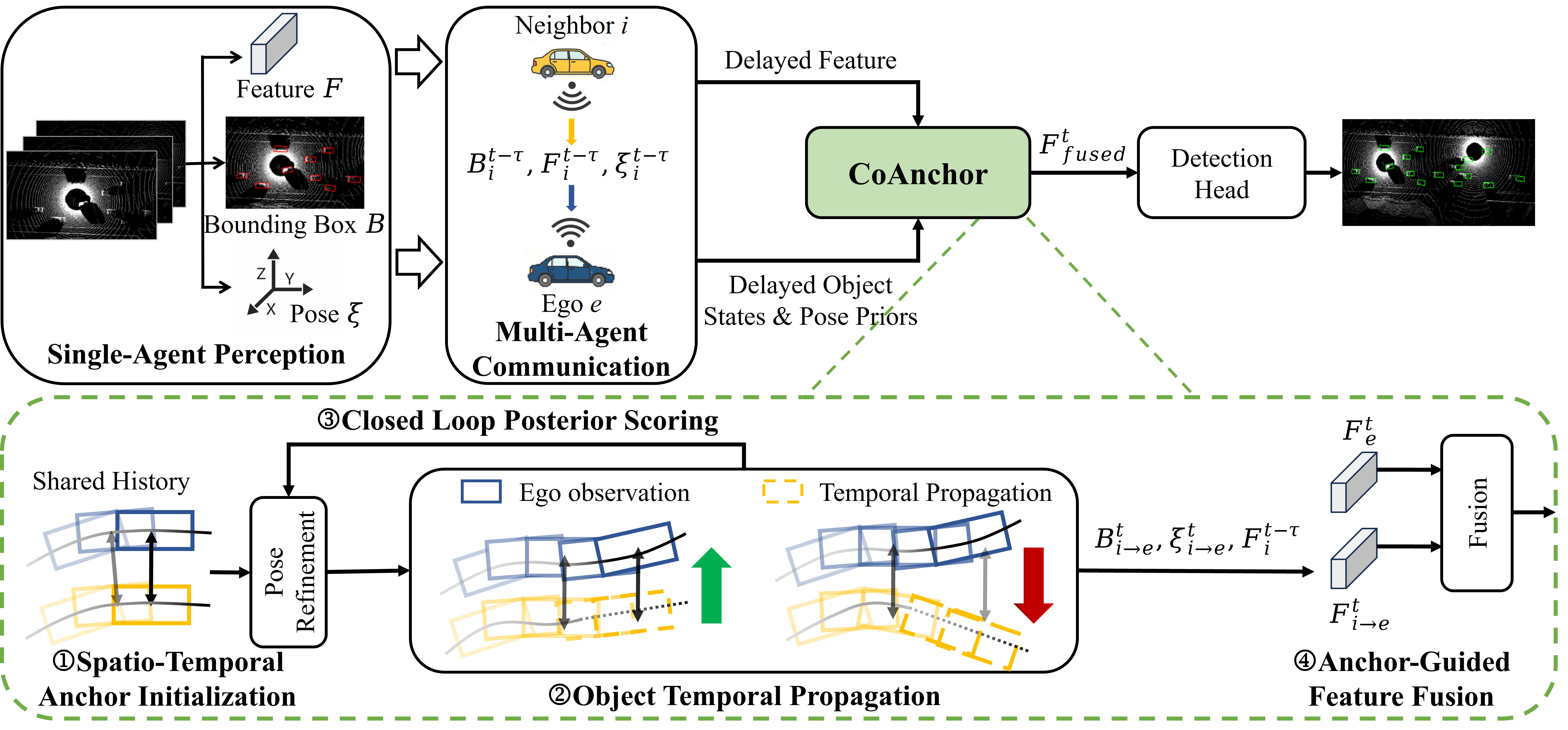}
		\Description{Pipeline of CoAnchor. Single-agent perception extracts BEV
		features, object boxes, and poses; multi-agent communication supplies
		delayed neighbor features and object states. CoAnchor performs
		spatio-temporal anchor initialization, object temporal propagation,
		closed-loop posterior scoring with pose refinement, and anchor-guided
		feature fusion before the detection head.}
	\vspace{-3mm}
	\caption{Proposed architecture of \name. Given delayed collaborator features, object cues, and noisy relative poses, \name constructs object-level anchors to refine the relative pose and to provide ego-guided priors for delay compensation. The propagated and verified anchor states are then fed back to later pose refinement and used to support subsequent collaborative fusion.\looseness=-1}
	\label{fig_overview}	
	\vspace{-4mm}
\end{figure*}

\subsection{Problem Formulation}
At ego time $t$, the message received from neighbor $j$ was captured earlier at the delayed time $u_j=t-\tau_j$, where $\tau_j$ denotes the communication delay. The relative pose used for collaboration is also noisy; for simplicity, we denote the delayed collaborator pose estimate by $\tilde{\xi}_{j}^{u_j}=\xi_{j}^{u_j}\oplus \epsilon_{j}^{u_j}$. In a direct collaborative pipeline, the delayed collaborator feature $F_j^{u_j}$ would be warped into the ego frame by the noisy relative pose and then fused with the ego feature $F_e^t$:
\begin{equation}
\mathcal{M}_{j\rightarrow e}^{u_j}
=
\phi_{\mathrm{warp}}\!\left(F_j^{u_j},\,T(\xi_e^t,\tilde{\xi}_j^{u_j})\right).
\label{eq_direct_pipeline}
\end{equation}
$T(\cdot)$ denotes the relative-pose transform from the collaborator frame to the ego frame and $\phi_{\mathrm{warp}}(\cdot)$ denotes BEV feature warping.
Under joint delay and pose noise, however, $\mathcal{M}_{j\rightarrow e}^{u_j}$ is not only spatially biased but temporally stale. Directly fusing it with $F_e^t$ can introduce shifted responses, duplicated structures, or motion ghosts in the final prediction. The goal of \name is to recover a reliable current-time collaborator representation before feature fusion.

\subsection{Overall Architecture} \label{sec_design_overview}

\name is an anchor-centric spatio-temporal alignment framework for collaborative perception under coupled communication delay and pose noise. It constructs a sparse object-level interface, because object states make cross-agent correspondence, motion evolution, and current-time verification much easier to control.

As shown in Fig.~\ref{fig_overview}, this sparse interface connects the whole pipeline. 
\emph{(i) Spatio-temporal anchor initialization} matches delayed neighbor objects with ego objects around the delayed timestamp, refines an initial relative pose, and converts the reliable matched pairs into delayed anchors. 
\emph{(ii) Object temporal propagation} advances delayed neighbor objects from $u_j$ to the ego current time $t$, while allowing anchor states to be corrected when a reliable current-time ego observation is available. 
\emph{(iii) Closed-loop posterior scoring} uses current-time agreement together with short-history consistency to update pair reliability, and feeds the updated weights back to the next pose-refinement round. 
\emph{(iv) Anchor-guided feature fusion} uses the final refined pose and propagated object hypotheses to correct collaborator features before downstream dense fusion and detection decoding.

\subsection{Spatio-Temporal Anchor Initialization} \label{sec_shared_object}

High-dimensional BEV features are informative for perception, but they do not explicitly expose object correspondences or temporal structure, especially under large delay and pose noise. 
By contrast, low-dimensional object states naturally carry spatial layout and motion information. We therefore start from object-level states and use them as the sparse carriers of spatio-temporal alignment.

\textbf{Object state and bipartite matching.} In \name's pipeline, each agent first runs its own single-agent detector to obtain the current object boxes, while the corresponding short box histories are retrieved from memory. For an object $m$ detected by agent $i$ at time $u$, we represent its motion state as
\begin{equation}
	x_{i,m}^{u}=[p_x,\,p_y,\,v_x,\,v_y,\,\psi]^T,
\end{equation}
where $(p_x,p_y)$ is the box center in the BEV plane, $(v_x,v_y)$ is the planar velocity estimated from short box histories, and $\psi$ is the heading angle. Given the delayed observations from neighbor $j$ at the delayed timestamp $u_j=t-\tau_j$, we first roughly warp the neighbor detections into the ego frame using the raw relative pose. 
We then construct cross-agent associations at $u_j$ in two stages. \emph{(i)} We apply a spatial-distance gate to discard obviously incompatible ego--neighbor pairs. \emph{(ii)} On the remaining candidates, we build a bipartite matching cost using velocity similarity together with local neighborhood consistency, and solve a one-to-one assignment by the Hungarian algorithm \cite{hungarian}. This yields the matched-pair set $\mathcal{P}_j=\{(a,b)\}$ between ego objects and delayed neighbor objects.

\textbf{Trajectory-guided pose refinement.} Each matched pair now provides not only a single-frame correspondence but a short motion history. We use these trajectory-aware pairs to refine the relative pose. For each matched pair $k=(a,b)\in\mathcal{P}_j$, we first convert its association cost into an initial reliability score $q_{j,k}^{(0)}$, so that pairs with lower matching cost start with higher confidence. At closed-loop round $\ell$, the refined transform $\Delta T_{j\to e}^{(\ell)}$ is obtained by iteratively reweighted least squares (IRLS):
\begingroup\small
\begin{equation}
\Delta T_{j\to e}^{(\ell)}
=
\arg\min_{\Delta T\in SE(2)}
\sum_{k=(a,b)\in\mathcal{P}_j}
q_{j,k}^{(\ell)}
\sum_{\kappa=0}^{L_p}
\rho\!\left(
\left\|
\Delta T\!\left(p_{j,b}^{u_j-\kappa}\right)-p_{e,a}^{u_j-\kappa}
\right\|_2^2
\right),
\label{eq_pose_irls}
\end{equation}
\endgroup
where $p_{j,b}^{u_j-\kappa}$ and $p_{e,a}^{u_j-\kappa}$ are the historical box centers of the neighbor object $b$ and ego object $a$ at offset $\kappa$, $L_p$ is the history length used in pose refinement, $q_{j,k}^{(\ell)}$ is the current reliability weight of pair $k$, and $\rho(\cdot)$ is a robust penalty. Intuitively, Eq.~\eqref{eq_pose_irls} searches for a single 2D rigid transform that best aligns the matched short trajectories.\looseness=-1

\textbf{Anchor construction.} After pose refinement, each matched pair is converted into a delayed spatio-temporal anchor. For pair $k$, we denote its delayed anchor state and covariance by $x_{j,k}^{u_j}$ and $\Sigma_{j,k}^{u_j}$, respectively. The initial pair score $q_{j,k}^{(0)}$, inherited from the association stage, is converted into $\Sigma_{j,k}^{u_j}$, so that low-confidence matches start with larger uncertainty before propagation. These anchors form the matched subset used for current-time correction and closed-loop feedback.

\subsection{Object Temporal Propagation} \label{sec_state_prop}
Next, the delayed neighbor scene must be advanced from the delayed timestamp $u_j$ to the current ego time $t$. We formulate this stage as prediction followed by optional current-time correction. Prediction uses only the delayed anchor state, while correction is applied only when a reliable current-time ego correspondence is available.
Each delayed neighbor box is first expressed in the ego coordinate frame at time $u_j$ using the refined relative pose. We then use the ego self-motion estimate to transform these delayed boxes into the ego coordinate frame at the current timestamp $t$, so that all later propagation, current-time correction, and feedback are performed in a unified current-time ego frame. 
After ego-time rewriting, we further propagate all delayed neighbor objects with a rule-based motion model, while reserving current-time correction only for the spatio-temporal anchors formed by matched pairs.

For each spatio-temporal anchor, we predict its current-time prior state and covariance as
\begin{equation}
x_{j,k}^{t-}=A(\tau_j)x_{j,k}^{u_j},
\label{eq_mu}
\end{equation}
\begin{equation}
\Sigma_{j,k}^{t-}=A(\tau_j)\Sigma_{j,k}^{u_j}A(\tau_j)^T+Q(\tau_j),
\label{eq_sigma}
\end{equation}
where $x_{j,k}^{u_j}$ and $\Sigma_{j,k}^{u_j}$ denote the delayed anchor state and covariance, $A(\tau_j)$ is the rule-based constant-velocity state-transition matrix over delay $\tau_j$ (with position updated by velocity and heading kept unchanged), and $Q(\tau_j)$ is the corresponding diagonal process-noise covariance that increases with delay.

For most propagated anchors, the ego still provides a reliable current-time observation. In such cases, we use the current ego observation of position and heading to perform a standard Kalman-style measurement update \cite{kalman} on the propagated prior. To determine whether the current-time correction should be accepted, we compute the normalized innovation squared
\begin{equation}
 d_{j,k}^{2}
 =
 (\nu_{j,k}^{t})^T (S_{j,k}^{t})^{-1}\nu_{j,k}^{t},
\end{equation}
where $\nu_{j,k}^{t}$ is the measurement residual and $S_{j,k}^{t}$ is its covariance. A small $d_{j,k}^{2}$ indicates that the propagated anchor agrees well with the current ego observation. If $d_{j,k}^{2}$ exceeds a $\chi^2$ gate, or if no reliable current-time ego correspondence is found due to missed detection, occlusion, or association failure, we do not maintain this hypothesis as an active anchor. Instead, it is downgraded to an ordinary propagated neighbor object.

This ordinary-object path also includes delayed neighbor objects that were not matched at initialization. All such objects are still advanced to time $t$ by the same ego-motion rewriting and rule-based motion model, so that the full delayed foreground can be brought into the current ego frame. However, since they are not maintained as spatio-temporal anchors, they do not participate in ego-side correction or later closed-loop feedback. In this way, only the subset that remains reliably supported by cross-time correspondence is preserved as active anchors for subsequent refinement. {When no reliable current-time ego correspondence is available, the collaborator object is retained through this ordinary propagation-and-fusion path rather than discarded. After gating, the next pose-refinement round is skipped if fewer than three valid trajectory correspondences remain.}

\subsection{Closed-Loop Posterior Scoring} \label{sec_closed_loop}

A matched pair is considered reliable only if it exhibits consistent spatio-temporal behavior over the available time span, as supported jointly by the shared pre-delay history and the current-time ego observation. We therefore use the propagated anchors to evaluate each pair before the next pose update.

For each matched pair with a reliable current-time ego correspondence, %
we first measure its current-time consistency by the normalized innovation $d_{j,k}^2$. We then check whether the pair remains stable over a small shared history window around the delayed timestamp, by comparing the pose-transformed neighbor boxes with the corresponding ego boxes across several nearby historical frames. This gives a short-history disagreement term $r_{j,k}^{\mathrm{hist}}$. We then combine the two into a unified feedback residual:
\begin{equation}
 r_{j,k}^{\mathrm{fb}} = d_{j,k}^{2} + r_{j,k}^{\mathrm{hist}}.
\end{equation}

The next-round pair weight is updated by preserving its initial confidence and exponentially downweighting it according to the feedback residual:
\begin{equation}
q_{j,k}^{(\ell+1)}
=
q_{j,k}^{(0)}
\exp\!\left(-\lambda_{\mathrm{fb}}\,r_{j,k}^{\mathrm{fb}}\right),
\end{equation}
where $q_{j,k}^{(0)}$ is the initial score of pair $k$, and $\lambda_{\mathrm{fb}}$ controls how strongly inconsistent pairs are suppressed. In this way, pairs that remain stable at both the current time and the nearby delayed-time history receive larger weights in the next IRLS pose update, while drifting correspondences are progressively downweighted.

\begin{table*}[t]
	\centering
	\footnotesize
	\setlength{\tabcolsep}{3.8pt}
	\renewcommand{\arraystretch}{1.06}
	\caption{Comparison with state-of-the-art methods on OPV2V and V2V4Real. Each entry is reported as AP@0.5 / AP@0.7.} %
	\label{tab:main_results}
	\vspace{-3mm}
	\begin{tabular}{l|cccc|cccc}
		\toprule
		\multirow{2}{*}{Method}
		& \multicolumn{4}{c|}{OPV2V \cite{opv2v}}
		& \multicolumn{4}{c}{V2V4Real \cite{v2v4real}} \\
		\cline{2-9}
		& No Noise & P-Hard & J-Hard & D-Hard
		& No Noise & P-Hard & J-Hard & D-Hard \\
		\hline
		CoAlign \cite{coalign} & 96.64 / 91.31 & 86.49 / 74.19 & 85.90 / 72.82 & 83.19 / 71.37 & 72.64 / 42.97 & 51.34 / 32.81 & 53.59 / 33.22 & 58.26 / 34.26 \\
		ERMVP \cite{ermvp} & 96.59 / \textbf{94.01} & 63.61 / 48.57 & 57.46 / 42.95 & 47.86 / 36.56 & 78.38 / \textbf{52.55} & 51.63 / 34.69 & 53.16 / 34.55 & 59.81 / 36.81 \\
		V2X-ViT \cite{v2xvit} & 95.71 / 87.54 & 77.34 / 60.06 & 75.90 / 55.94 & 68.55 / 52.28 & 78.33 / 51.45 & 51.87 / 35.37 & 53.83 / 36.24 & 59.84 / 37.62 \\
		CoST \cite{cost} & 97.20 / 93.70 & 51.00 / 32.10 & 47.90 / 27.80 & 42.60 / 26.90 & 74.10 / 50.00 & 49.50 / 30.60 & 51.80 / 32.00 & 58.00 / 35.80 \\
		TraF-Align (ego-only) & 84.49 / 78.57 & 84.49 / 78.57 & 84.49 / 78.57 & 84.49 / 78.57 & 62.19 / 40.55 & 62.19 / 40.55 & 62.19 / 40.55 & 62.19 / 40.55 \\
		TraF-Align \cite{traf_align} & \textbf{97.22} / 93.66 & 82.27 / 69.43 & 85.87 / 72.46 & 90.33 / 78.58 & \textbf{80.60} / 52.29 & 55.33 / 35.30 & 58.18 / 35.09 & 62.79 / 37.47 \\
		\base \cite{coalign,traf_align} & 97.04 / 92.90 & 85.97 / 76.34 & 90.19 / 81.21 & 93.76 / 86.03 & 74.82 / 44.89 & 56.30 / 36.15 & 61.04 / 37.69 & 66.23 / 40.86 \\
		\name (ego-only) & 83.09 / 72.35 & 83.09 / 72.35 & 83.09 / 72.35 & 83.09 / 72.35 & 59.93 / 40.43 & 59.93 / 40.43 & 59.93 / 40.43 & 59.93 / 40.43 \\
		\hline
		\textbf{\name}        & 96.89 / 92.08 & \textbf{95.85} / \textbf{89.36} & \textbf{95.15} / \textbf{87.96} & \textbf{94.44} / \textbf{86.40} & 77.71 / 50.56 & \textbf{67.76} / \textbf{45.01} & \textbf{67.04} / \textbf{44.05} & \textbf{66.68} / \textbf{43.50} \\
		\bottomrule
	\end{tabular}
	\vspace{-4mm}
\end{table*}

\subsection{{}{Anchor-Guided Feature Fusion}} \label{sec_anchor_fusion}

After the closed-loop stage, we retain the final refined relative pose together with the propagated current-time object hypotheses. The delayed high-dimensional neighbor feature is first rewritten into the ego current frame using the refined pose:
\begin{equation}
\mathcal{M}_{j\rightarrow e}^{u_j}
=
\phi_{\mathrm{warp}}
\Big(
F_j^{u_j},
T({}{\xi_e^t},\tilde{\xi}_j^{u_j})
\circ
\Delta T_{j\rightarrow e}^{(\ell_{\max})}
\Big),
\end{equation}
where $\Delta T_{j\rightarrow e}^{(\ell_{\max})}$ is the final refined pose after the last loop round. 
Specifically, the delayed collaborator feature is first transformed into the ego coordinate system at the delayed time, and then brought into the ego frame at the current time.
We apply a non-learnable box-wise feature mover: for each delayed neighbor box already rewritten into the current ego frame, it extracts the enclosing axis-aligned BEV rectangle and relocates it toward the corresponding propagated current-time box by bilinear sampling. This deterministic operation corrects the dominant foreground displacement before dense fusion, while leaving the rest of the BEV feature unchanged.

\textbf{Downstream fusion and detection decoding.} After pose correction and box-wise foreground adjustment, the neighbor features are aggregated with the ego feature by a downstream collaborative perception module. The fusion operator here is flexible and can be instantiated by any standard cooperative perception module. In our implementation, we adopt multi-scale feature fusion, since it achieves a good trade-off among computational cost, optimization stability, and multi-scale representation capability, thus providing robust detection performance in practice. The fused feature maps are then decoded into final detection outputs. Concretely, the decoder produces a regression output and a classification output for candidate boxes. The regression output describes the object geometry, including object position, box size, and heading angle, while the classification output estimates the confidence that each candidate belongs to a foreground vehicle rather than background. These predictions are finally decoded to obtain the detection results.

\section{Evaluation} \label{sec_eval}

\subsection{Experimental Setup} \label{sec_eval_setup}

\textbf{Datasets.}
We evaluate the proposed method on two widely used collaborative perception benchmarks, OPV2V \cite{opv2v} and V2V4Real \cite{v2v4real}. OPV2V is a large-scale simulated benchmark for vehicle-to-vehicle collaboration, while V2V4Real is a real-world benchmark collected in realistic driving scenes. %

\textbf{Compared methods.}
We compare our method with representative baselines from both pose-robust and asynchronous collaborative perception, including V2X-ViT \cite{v2xvit}, ERMVP \cite{ermvp}, CoAlign \cite{coalign}, TraF-Align \cite{traf_align}, and CoST \cite{cost}. To verify whether spatial refinement and delayed fusion can be coordinated by simple composition, we further construct the cascaded baseline \base. We also report ego-only references for TraF-Align and our method to reveal how much collaborative gain remains under heavy perturbations.

\textbf{Implementation details.}
For a fair comparison, all methods adopt {}{PointPillars~\cite{pointpillars}} as the common BEV backbone. Following standard collaborative perception settings, the voxel size is set to 0.4\,m and the communication range is limited to 70\,m. We adopt a staged training pipeline. We first train a single-agent detector to produce per-agent object detections. During collaborative training, these detections are used as the object-level inputs of CoAnchor, and the whole collaborative perception model is trained under injected pose noise together with the downstream collaborative detection branch.  The collaborative detector is supervised by the standard classification, box-regression, and direction losses used in {}{PointPillars-style} heads. All models are trained within the same codebase on two NVIDIA RTX 4090 GPUs. We follow the official optimization settings of each baseline whenever available. {}

\textbf{Evaluation protocol.}
We report detection performance using AP@0.5 and AP@0.7. To evaluate robustness under coupled perturbations, we inject Gaussian pose noise into each agent pose in the BEV plane and impose communication delay on the non-ego branch by delaying collaborator observations. Specifically, for a pose $\xi=(x,y,\theta)$, the noisy pose is constructed by independently adding noise with standard deviation $\sigma_t$ to $x$ and $y$, and noise with standard deviation $\sigma_r$ to $\theta$. Unless otherwise specified, pose noise is written as $(\sigma_t,\sigma_r)$, where $\sigma_t$ is measured in meters and $\sigma_r$ in degrees.
Our main comparison is conducted under four representative settings: \emph{(i) No Noise}, with no injected pose perturbation or communication delay; \emph{(ii) Joint-Hard}, with 200\,ms delay and pose noise $(0.6\,\text{m}, 0.6^\circ)$; \emph{(iii) Pose-Hard}, with 100\,ms delay and pose noise $(0.9\,\text{m}, 0.9^\circ)$; and \emph{(iv) Delay-Hard}, with 300\,ms delay and pose noise $(0.3\,\text{m}, 0.3^\circ)$. Unless otherwise specified, the same fixed rule-based settings are used across datasets and test conditions.

\begin{figure*}[t]
		\centering
		\small
		\Description{Four line charts report detection average precision as
		perturbations increase. The top row shows OPV2V AP at IoU thresholds 0.5
		and 0.7 as pose noise grows under a fixed 100 millisecond delay. The
		bottom row shows V2V4Real AP at the same thresholds as delay grows under
		fixed pose noise. CoAnchor remains the highest and most stable curve,
		while most baselines decline more sharply.}
		\begin{minipage}[t]{0.24\textwidth}
		\centering
		\includegraphics[width=\linewidth]{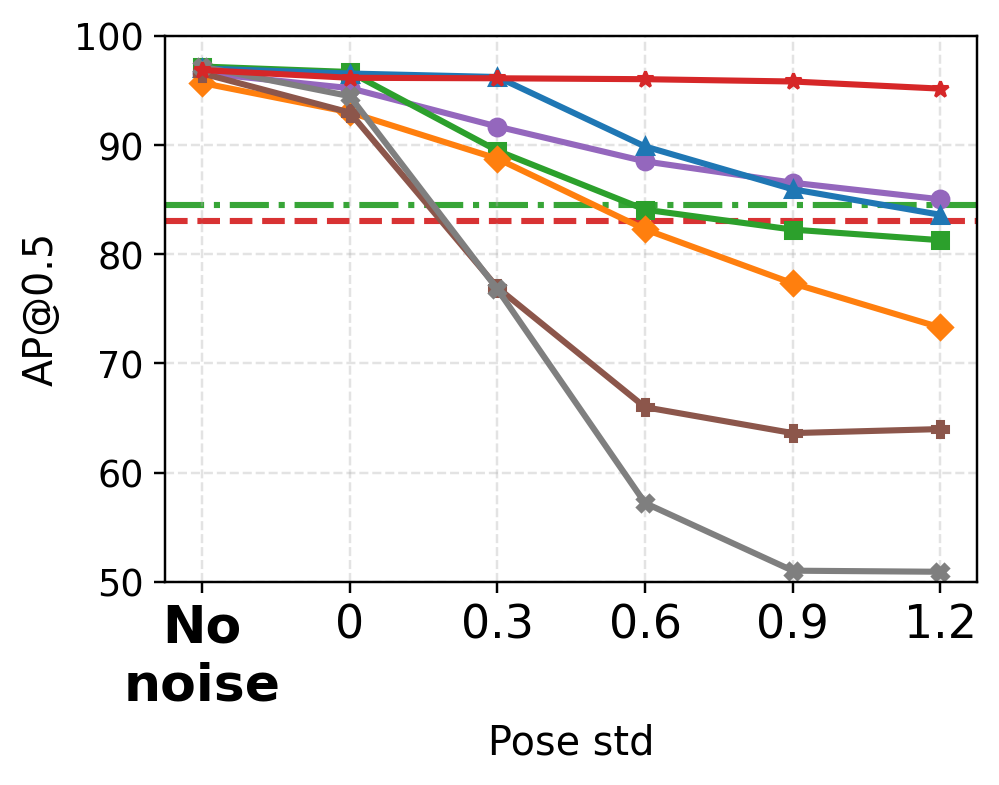}
		\vspace{-1mm}
		{\hspace*{0.13\linewidth}\parbox[t]{0.87\linewidth}{\centering (a) OPV2V, fixed delay = {}{100\,ms}}}
	\end{minipage}\hfill
	\begin{minipage}[t]{0.24\textwidth}
		\centering
		\includegraphics[width=\linewidth]{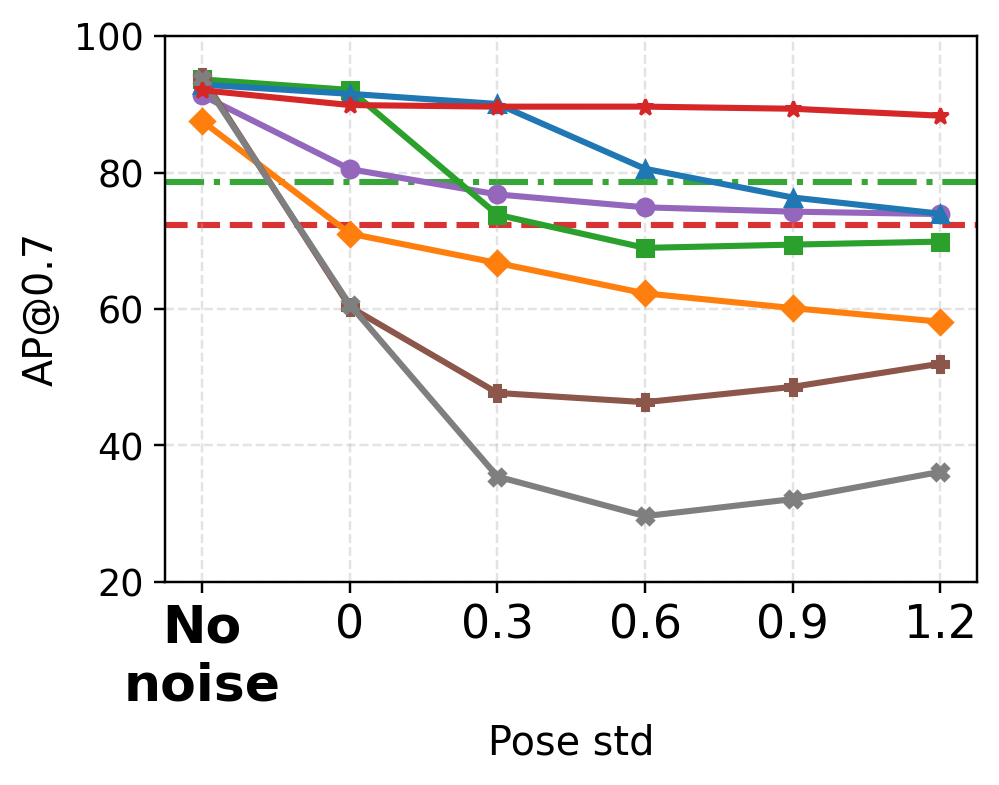}
		\vspace{-1mm}
		{\hspace*{0.13\linewidth}\parbox[t]{0.87\linewidth}{\centering (b) OPV2V, fixed delay = {}{100\,ms}}}
	\end{minipage}\hfill
	\begin{minipage}[t]{0.24\textwidth}
		\centering
		\includegraphics[width=\linewidth]{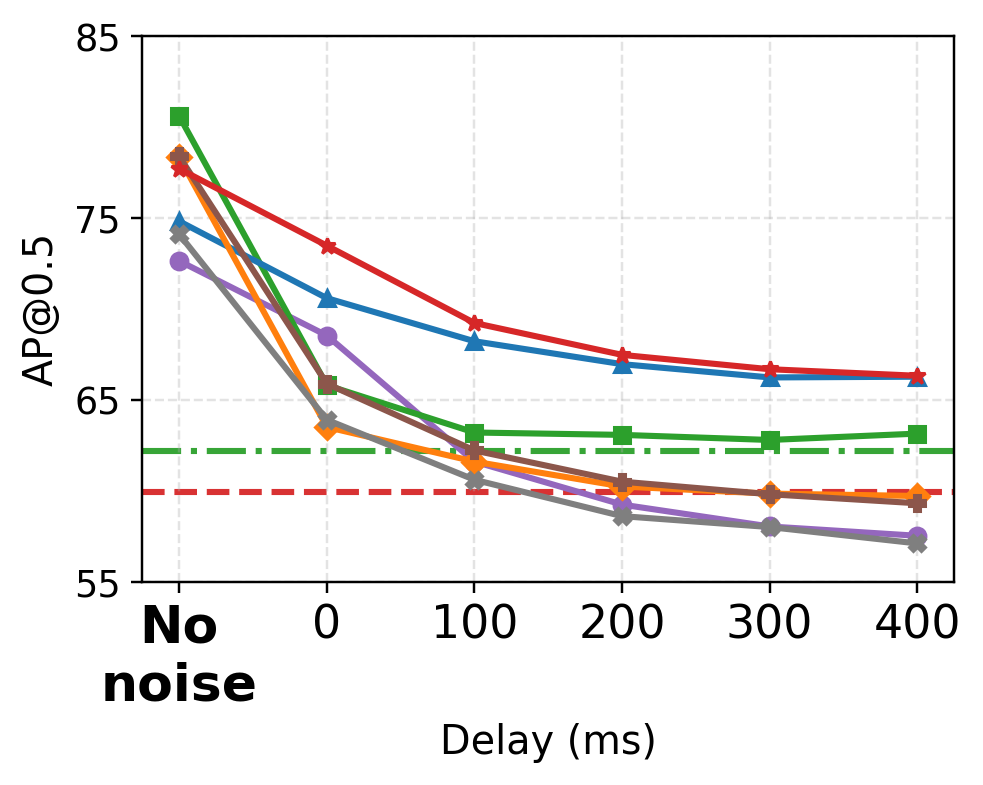}
		\vspace{-1mm}
		{\hspace*{0.10\linewidth}\parbox[t]{0.90\linewidth}{\centering (c) V2V4Real, fixed pose std = {}{$(0.3\,\mathrm{m}, 0.3^\circ)$}}}
	\end{minipage}\hfill
	\begin{minipage}[t]{0.24\textwidth}
		\centering
		\includegraphics[width=\linewidth]{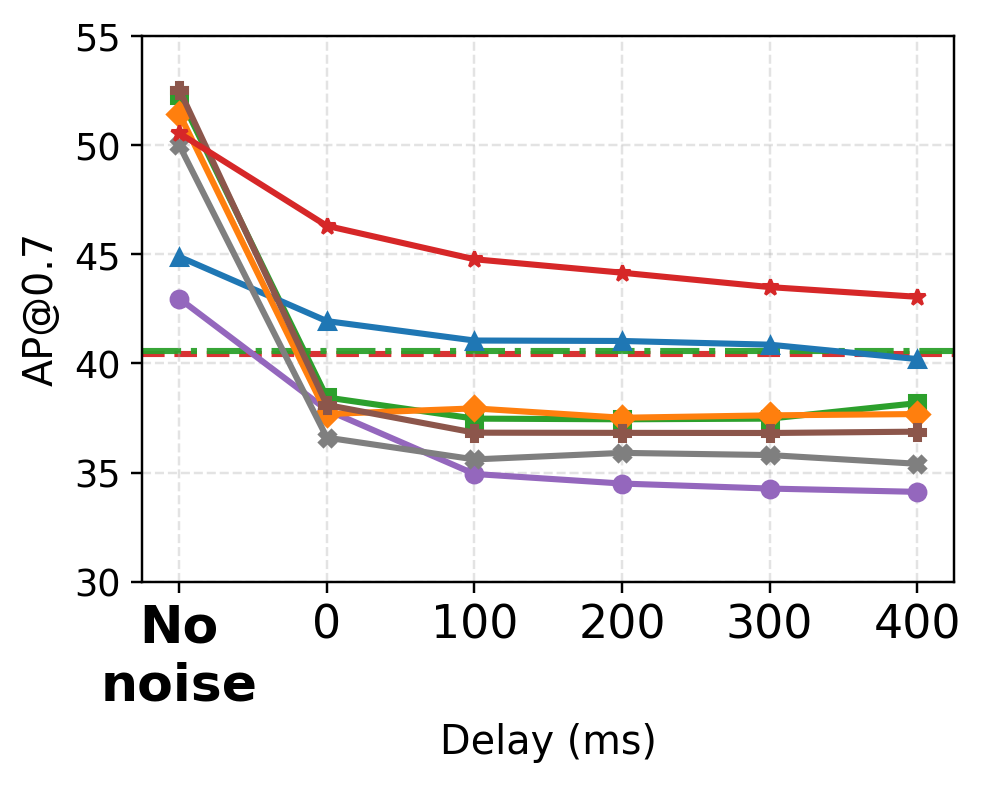}
		\vspace{-1mm}
		{\hspace*{0.10\linewidth}\parbox[t]{0.90\linewidth}{\centering (d) V2V4Real, fixed pose std = {}{$(0.3\,\mathrm{m}, 0.3^\circ)$}}}
	\end{minipage}
	\vspace{1mm}
	{\small
		\setlength{\tabcolsep}{0pt}
		\renewcommand{\arraystretch}{1.05}
		\hspace*{0.01\textwidth}\begin{tabular*}{0.98\textwidth}{@{\extracolsep{\fill}}>{\raggedright\arraybackslash}p{0.18\textwidth}>{\raggedright\arraybackslash}p{0.18\textwidth}>{\raggedright\arraybackslash}p{0.18\textwidth}>{\raggedright\arraybackslash}p{0.18\textwidth}>{\raggedright\arraybackslash}p{0.18\textwidth}@{}}
			\makebox[2.9em][l]{\raisebox{-0.7ex}{\tikz[baseline=-0.6ex]{
						\draw[red!80!black,line width=1.1pt] (0,0.09) -- (0.34,0.09);
						\draw[red!80!black,line width=1.1pt] (0.17,0.03) -- (0.17,0.15);
						\draw[red!80!black,line width=1.1pt] (0.11,0.09) -- (0.23,0.09);
						\draw[red!80!black,line width=1.1pt] (0.125,0.045) -- (0.215,0.135);
						\draw[red!80!black,line width=1.1pt] (0.125,0.135) -- (0.215,0.045);
			}}}\name &
			\makebox[2.9em][l]{\raisebox{-0.7ex}{\tikz[baseline=-0.6ex]{
						\draw[green!60!black,line width=1.1pt] (0,0.09) -- (0.34,0.09);
						\fill[green!60!black] (0.13,0.05) rectangle (0.21,0.13);
			}}}TraF-Align &
			\makebox[2.9em][l]{\raisebox{-0.7ex}{\tikz[baseline=-0.6ex]{
						\draw[blue!70!black,line width=1.1pt] (0,0.09) -- (0.34,0.09);
						\fill[blue!70!black] (0.17,0.15) -- (0.11,0.03) -- (0.23,0.03) -- cycle;
			}}}\base &
			\makebox[2.9em][l]{\raisebox{-0.7ex}{\tikz[baseline=-0.6ex]{
						\draw[orange!90!black,line width=1.1pt] (0,0.09) -- (0.34,0.09);
						\fill[orange!90!black] (0.17,0.15) -- (0.23,0.09) -- (0.17,0.03) -- (0.11,0.09) -- cycle;
			}}}V2X-ViT &
			\\
		\end{tabular*}\par\vspace{-1.2ex}
		\noindent\hspace*{0.017\textwidth}\begin{tabular*}{0.98\textwidth}{@{\extracolsep{\fill}}>{\raggedright\arraybackslash}p{0.18\textwidth}>{\raggedright\arraybackslash}p{0.18\textwidth}>{\raggedright\arraybackslash}p{0.18\textwidth}>{\raggedright\arraybackslash}p{0.18\textwidth}>{\raggedright\arraybackslash}p{0.18\textwidth}@{}}
			\makebox[2.9em][l]{\raisebox{-0.7ex}{\tikz[baseline=-0.6ex]{
						\draw[brown!70!black,line width=1.1pt] (0,0.09) -- (0.34,0.09);
						\draw[brown!70!black,line width=1.1pt] (0.17,0.03) -- (0.17,0.15);
						\draw[brown!70!black,line width=1.1pt] (0.11,0.09) -- (0.23,0.09);
			}}}ERMVP &
			\makebox[2.9em][l]{\raisebox{-0.7ex}{\tikz[baseline=-0.6ex]{
						\draw[gray!70!black,line width=1.1pt] (0,0.09) -- (0.34,0.09);
						\draw[gray!70!black,line width=1.1pt] (0.12,0.04) -- (0.22,0.14);
						\draw[gray!70!black,line width=1.1pt] (0.12,0.14) -- (0.22,0.04);
			}}}CoST &
			\makebox[2.9em][l]{\raisebox{-0.7ex}{\tikz[baseline=-0.6ex]{
						\draw[draw={rgb,255:red,148;green,103;blue,189},line width=1.1pt] (0,0.09) -- (0.34,0.09);
						\fill[fill={rgb,255:red,148;green,103;blue,189}] (0.17,0.09) circle (0.04);
			}}}CoAlign &
			\makebox[2.9em][l]{\raisebox{-0.7ex}{\tikz[baseline=-0.6ex]{
						\draw[red!80!black,dashed,line width=1.1pt] (0,0.09) -- (0.34,0.09);
			}}}\name ego-only &
			\makebox[2.9em][l]{\raisebox{-0.7ex}{\tikz[baseline=-0.6ex]{
						\draw[green!60!black,dash dot,line width=1.1pt] (0,0.09) -- (0.34,0.09);
			}}}TraF ego-only
			\\
	\end{tabular*}}
	\vspace{-4mm}
	\caption{Robustness analysis under progressively increasing temporal and spatial perturbations. Panels (a)--(b) fix the delay at {}{100\,ms} on OPV2V and vary the pose-noise magnitude. Panels (c)--(d) {}{fix the pose-noise standard deviations at $(0.3\,\mathrm{m}, 0.3^\circ)$} on V2V4Real and vary the communication delay. AP@0.5 and AP@0.7 are reported separately.}
	\label{fig:robustness_curves}
	\vspace{-4mm}
\end{figure*}

\subsection{Quantitative Comparison} %

\textbf{Main results.}
Table~\ref{tab:main_results} reports the comparison with state-of-the-art methods. Overall, \name is not always the best under the clean setting, but it remains competitive and {}{achieves the strongest robustness} once pose noise and communication delay are jointly introduced. This trend is consistent across both OPV2V and V2V4Real, indicating that the proposed \name preserves standard-case accuracy while substantially improving robustness under coupled perturbations.
\emph{(i)} On OPV2V, our method reaches 95.15 / 87.96 under \emph{Joint-Hard}, clearly outperforming the strongest baseline \base at 90.19 / 81.21. The advantage is already visible in the more pose-dominant mixed setting \emph{Pose-Hard}, while under \emph{Delay-Hard} our method still remains slightly above \base.
\emph{(ii)} On V2V4Real, where the collaborative pipeline is more strongly affected by realistic detection noise and unstable cross-agent correspondences, the robustness gain becomes even more pronounced. Under \emph{Joint-Hard}, \name reaches 67.04 / 44.05, clearly surpassing \base at 61.04 / 37.69. Under \emph{Pose-Hard}, the margin is even larger, and under \emph{Delay-Hard} our method still maintains an advantage, especially on AP@0.7.

Moreover, four observations are worth highlighting. \textit{(i)} Data-driven fusion methods such as V2X-ViT remain competitive in the clean setting, but their performance drops sharply once the perturbation becomes large, especially on the harder mixed and joint settings. This suggests that implicit feature-level fitting has limited extrapolation ability when the test-time noise exceeds the regime that can be absorbed by the learned alignment. \textit{(ii)} {}{Spatially oriented} methods such as CoAlign can benefit from explicit pose handling, but their gains shrink once delayed observations also need to be propagated to the current ego time. This is particularly clear on V2V4Real, where their clean-case accuracy is already noticeably below strong asynchronous baselines. \textit{(iii)} The delay-oriented method TraF-Align remains very strong when the cross-agent spatial reference is reliable, but its robustness becomes sensitive to pose corruption. In both datasets, its clean or mildly perturbed performance is high, whereas the gap to our method enlarges in the harder coupled settings. \textit{(iv)} The cascaded baseline \base is indeed stronger than single-purpose modules in several noisy regimes, which confirms that simple composition is useful but insufficient: it still leaves a clear gap to \name under \emph{Joint-Hard}. This behavior matches our design goal: instead of only stacking spatial and temporal corrections, we explicitly verify which propagated object hypotheses remain trustworthy at the current ego time and feed this reliability back to later pose refinement.

\textbf{Robustness to increasing pose noise.}
{}
{We fix the delay and gradually increase pose noise on OPV2V, as shown in Figure~\ref{fig:robustness_curves} (a) and (b). At low noise, \base remains competitive because temporal compensation is effective while the upstream spatial reference is still reliable. As pose noise grows, V2X-ViT and \base degrade more rapidly: learned feature interaction has limited tolerance to large perturbations, while temporal propagation alone cannot determine whether incoming collaborator evidence is already spatially biased. In contrast, \name combines short-history pose refinement with current-time anchor verification, preserving valid hypotheses while suppressing spatially biased ones and therefore degrading more gracefully across the tested noise range.}

\textbf{Robustness to increasing delay.}
{}
{We fix the pose noise and vary the communication delay on V2V4Real, as shown in Figure~\ref{fig:robustness_curves} (c) and (d). TraF-Align drops sharply once the collaborator input is spatially biased, indicating that temporal compensation starts from an unreliable spatial reference under coupled perturbations. \base degrades more slowly as the delay increases because upstream pose correction provides partial mitigation, but it starts from a noticeably lower level, consistent with negative pose optimization under realistic detection noise. By re-evaluating propagated anchors with current-time ego observations before later pose refinement and fusion, \name retains valid collaborator hypotheses while filtering stale or spatially biased ones, maintaining the strongest and most stable collaborative advantage across the tested delay range.}\looseness=-1
\begin{figure}
		\centering
		\Description{Four LiDAR bird's-eye-view detection maps compare CoAlign,
		V2X-ViT, baseST, and CoAnchor under the Joint-Hard setting. Yellow
		callouts enlarge crowded regions. CoAnchor shows closer overlap between
		the red and green boxes and fewer visibly shifted or duplicated
		detections than the comparison methods.}
			\begin{minipage}[t]{0.48\linewidth}
			\centering
			\includegraphics[width=0.92\linewidth,height=0.59\linewidth]{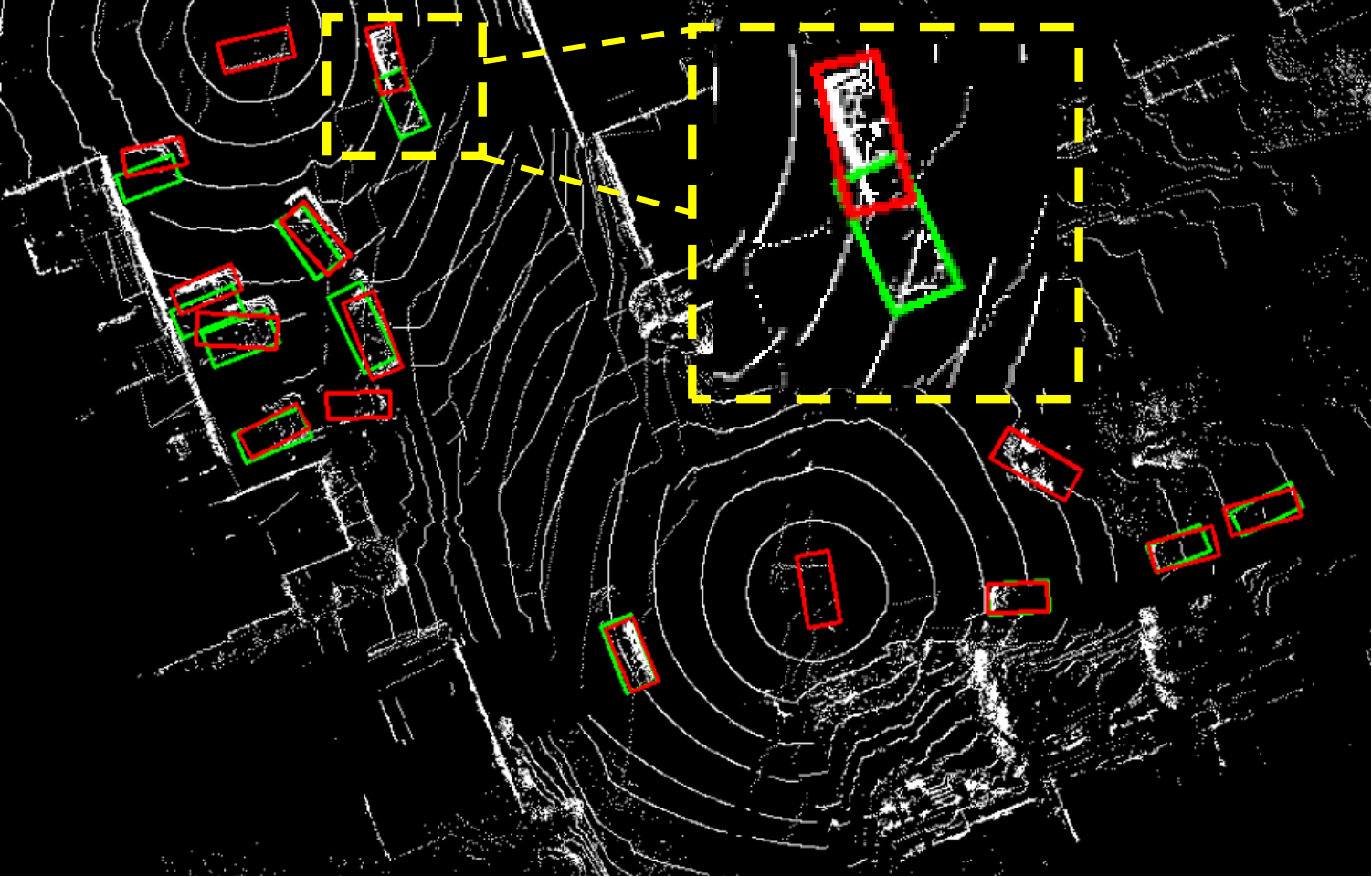}
			\vspace{-4mm}
			\caption*{(a) CoAlign}
		\end{minipage}\hfill
		\begin{minipage}[t]{0.48\linewidth}
			\centering
			\includegraphics[width=0.92\linewidth,height=0.59\linewidth]{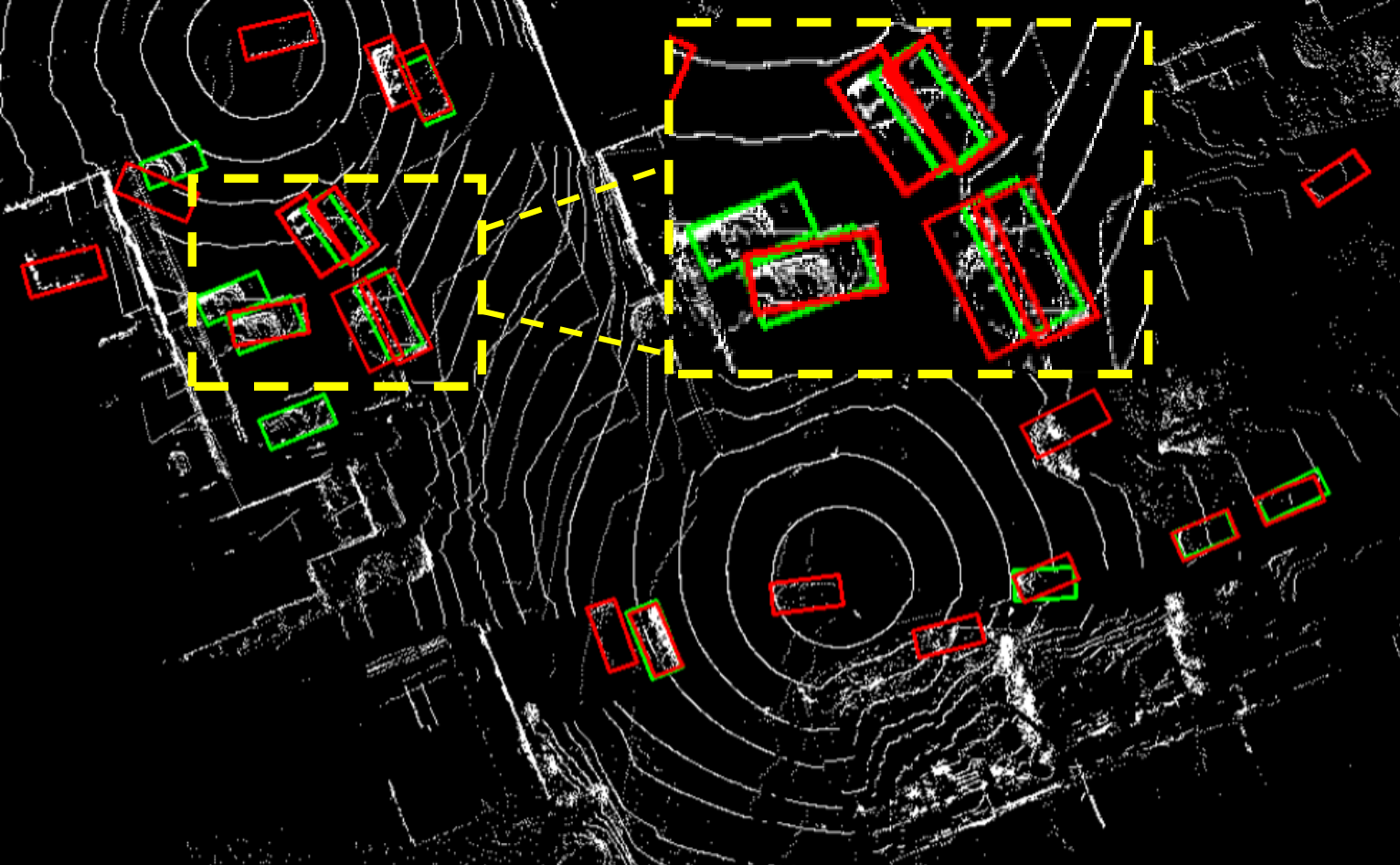}
			\vspace{-4mm}
			\caption*{(b) V2X-ViT}
		\end{minipage}
		\begin{minipage}[t]{0.48\linewidth}
			\centering
			\includegraphics[width=0.92\linewidth,height=0.59\linewidth]{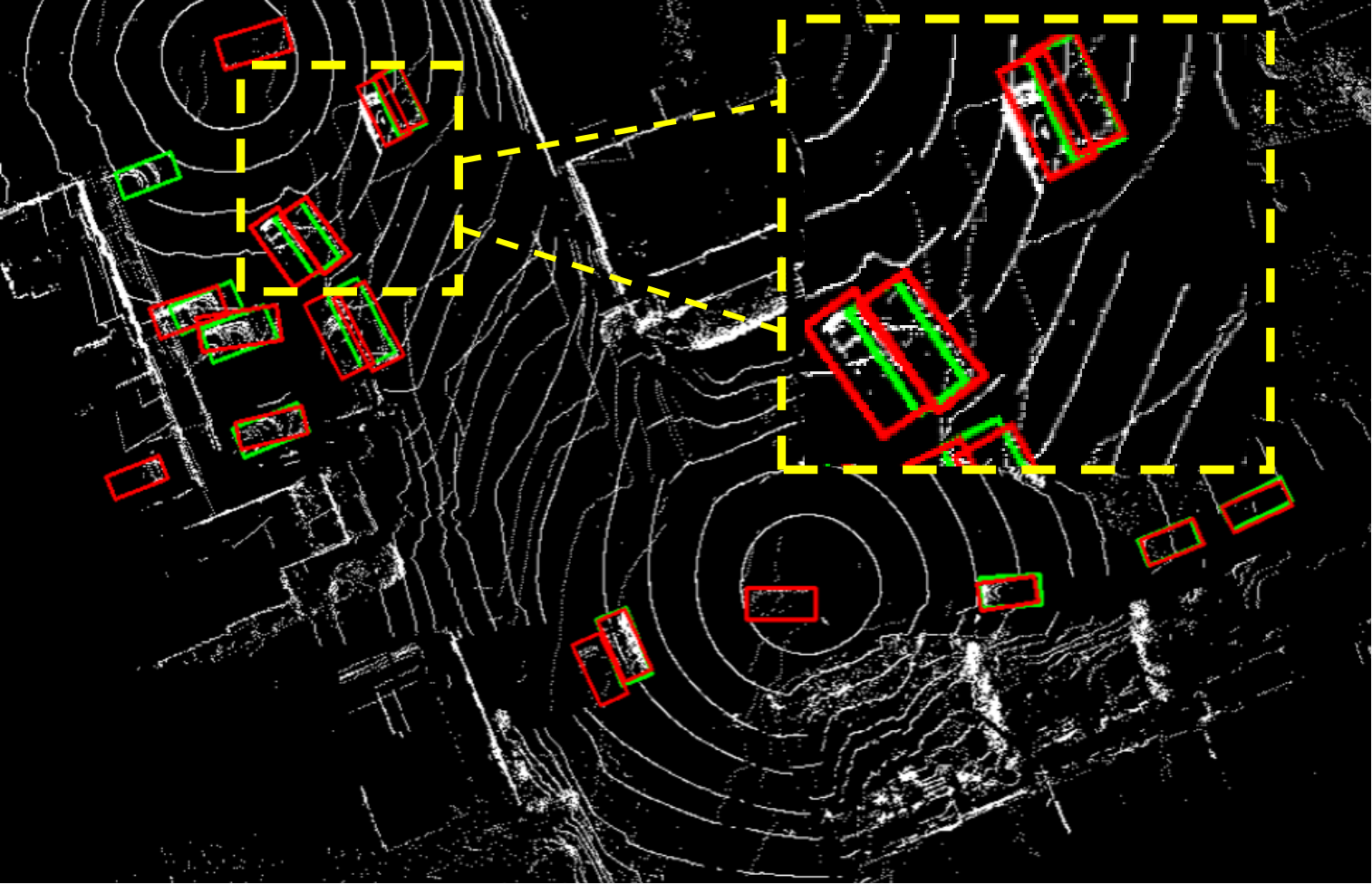}
			\vspace{-4mm}
			\caption*{(c) \base}
		\end{minipage}\hfill
		\begin{minipage}[t]{0.48\linewidth}
			\centering
			\includegraphics[width=0.92\linewidth,height=0.59\linewidth]{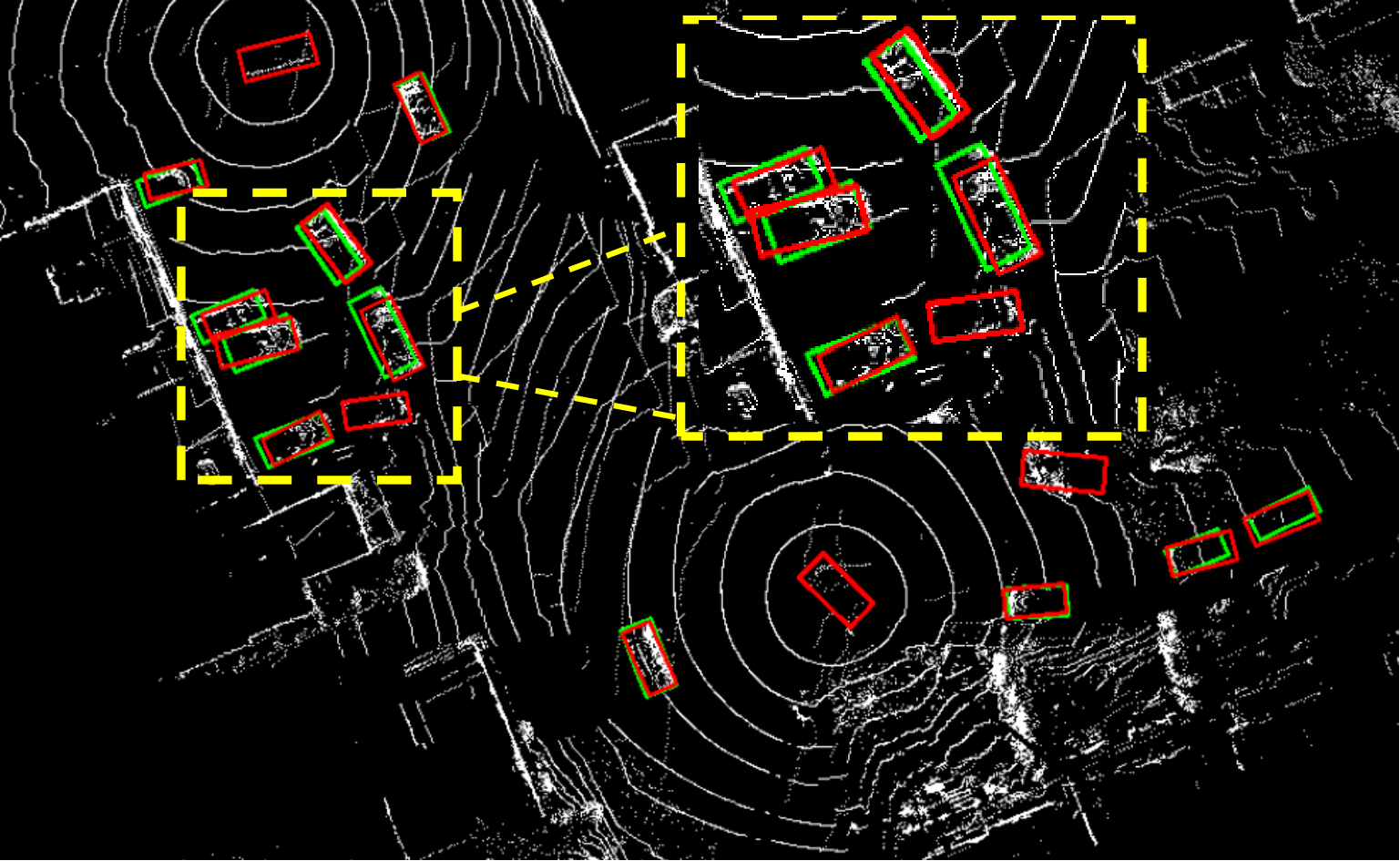}
			\vspace{-4mm}
			\caption*{(d) \name}
		\end{minipage}
	\vspace{-4mm}
	\caption{Qualitative analysis under the \emph{Joint-Hard} setting.}
	\vspace{-4mm}
	\label{fig:qualitative_results}
\end{figure}

\subsection{Qualitative Comparison}

Figure~\ref{fig:qualitative_results} shows representative qualitative results under coupled pose noise and communication delay. \base tends to produce duplicate responses together with missed detections when delayed collaborator evidence is not properly verified before fusion. CoAlign alleviates part of the spatial misalignment, but under asynchronous inputs, it can still leave delayed boxes caused by the temporal gap, and some collaborator boxes remain spatially shifted after pose correction. V2X-ViT exhibits another typical failure pattern, where missed detections and duplicate responses coexist once the learned fusion is affected by stronger spatio-temporal perturbation.

By comparison, \name produces cleaner final detections with fewer duplicates and more accurate box locations. {} During object propagation, {}{ego-guided correction} helps merge propagated co-visible hypotheses with the current ego observations and suppress duplicate ghost responses before final fusion. The closed-loop pose feedback further removes unreliable matched pairs in later pose refinement rounds, reducing the chance that spatially biased or temporally stale collaborator evidence survives into the final fused detections. Overall, the qualitative behavior is consistent with the quantitative results: the proposed framework better preserves useful collaborator information while suppressing duplicated or shifted responses under coupled perturbations.

\subsection{Ablation Study}
{}

{}

{}

{\textbf{Module-wise ablation.}
Table~\ref{tab:ablation_modules} evaluates three anchor-level components on OPV2V under the Joint-Hard setting. Specifically, {w/o historical initialization cues} removes short-history information during anchor initialization; {w/o ego-guided correction} prevents propagated anchors from being corrected by reliable current-time ego observations; and {w/o anchor feedback} disables current-time reliability updates before later pose-refinement rounds.}

{Removing any component degrades performance. Removing historical initialization cues causes the largest drop, from $95.15/87.96$ to $92.47/81.59$. Removing ego-guided correction yields $94.03/85.27$, while removing anchor feedback gives $94.31/86.73$. These results show that short-history cues establish reliable initial anchors, while current-time ego correction and anchor feedback provide complementary gains beyond a feed-forward anchor pipeline.}

\begin{table}
	\centering
	\small
	\caption{Module-wise ablation study on OPV2V.} %
	\label{tab:ablation_modules}
	\vspace{-4mm}
	\begin{tabular}{l|cc}
		\toprule
		\multirow{2}{*}{Variant}
		& \multicolumn{2}{c}{OPV2V} \\
		\cline{2-3}
		& AP@0.5 & AP@0.7 \\
		\midrule
		w/o historical initialization cues  & 92.47 & 81.59 \\
		w/o ego-guided correction  & 94.03 & 85.27 \\
		w/o anchor feedback  & 94.31 & 86.73 \\
		\hline
		\textbf{\name (Full)}      & \textbf{95.15} & \textbf{87.96} \\
		\bottomrule
	\end{tabular}
	\vspace{-4mm}
\end{table}

\begin{table}
	\centering
	\small
	\setlength{\tabcolsep}{5pt}
	\caption{Performance and runtime analysis on V2V4Real.} %
	\label{tab:ablation_runtime}
	\vspace{-4mm}
	\begin{tabular}{l|cc|c}
		\toprule
		\multirow{2}{*}{Method}
		& \multicolumn{2}{c|}{V2V4Real}
		& \multirow{2}{*}{\shortstack{Time\\(ms/frame)}} \\
		& AP@0.5 & AP@0.7 & \\
		\midrule
		V2X-ViT \cite{v2xvit}                    & 53.83 & 36.24 & 110.62 \\
		TraF-Align \cite{traf_align}           & 58.18 & 35.09 & 121.73 \\
		\name w/o feedback & 66.19 & 43.35 & 49.39 \\
		\hline
		\name +1 round feedback         & \textbf{67.04} & 44.05 & 51.62 \\
		\name +2 round feedback        & 66.96 & \textbf{44.16} & 52.51 \\
		\bottomrule
	\end{tabular}
	\vspace{-6mm}
\end{table}

\textbf{Efficiency and runtime analysis.}
{}
{We further compare detection accuracy and runtime on V2V4Real under the \emph{Joint-Hard} setting in Table~\ref{tab:ablation_runtime}. We include the external baselines {V2X-ViT} and {TraF-Align}, together with three versions of our anchor-centric pipeline: {\name w/o feedback}, {\name +1 round feedback}, and {\name +2 round feedback}. Because the closed loop operates mainly on sparse anchors rather than repeatedly invoking dense fusion modules, additional feedback incurs only limited computational overhead.} {} Compared with {\name w/o feedback}, adding one feedback round improves accuracy with only a small runtime increase, indicating that a single anchor-feedback step already brings the main practical gain. Adding a second round yields only marginal change, so most of the benefit is already obtained in the first round.
It is also worth noting that our method remains substantially more efficient than the compared baselines. Even with one feedback round, it is still much faster than V2X-ViT and TraF-Align while achieving the best detection accuracy in this setting. Therefore, one feedback round offers the most favorable accuracy--efficiency trade-off and is adopted as the default setting. %

\section{CONCLUSION AND FUTURE WORK}

In this paper, we study asynchronous collaborative perception under coupled communication delay and relative-pose noise. We propose \name, an anchor-centric closed-loop spatio-temporal alignment framework that uses sparse object-level anchors to connect short-history-aware initialization, ego-guided propagation, current-time feedback, and pose-corrected feature fusion. {}{Experiments show that \name remains competitive under clean settings while improving robustness under coupled perturbations. Although the adopted constant-velocity motion model provides a lightweight approximation for short communication delays, its modeling capacity may be limited under longer latency or highly nonlinear object motion. Future work will explore stronger delay-aware motion models and improve standard-case accuracy while preserving the anchor-centric design's efficiency and robustness.}{}

\begin{acks}
We appreciate the insightful feedback from the anonymous reviewers who helped
improve this work. This work was supported by the NSFC Project (62402058), the
Foundation of State Key Laboratory of Networking and Switching Technology
(NST20260111), and the Fundamental Research Funds for the Central Universities.
\end{acks}

\bibliographystyle{ACM-Reference-Format}
\bibliography{sample-base}


\begin{thebibliography}{37}


\ifx \showCODEN    \undefined \def \showCODEN     #1{\unskip}     \fi
\ifx \showISBNx    \undefined \def \showISBNx     #1{\unskip}     \fi
\ifx \showISBNxiii \undefined \def \showISBNxiii  #1{\unskip}     \fi
\ifx \showISSN     \undefined \def \showISSN      #1{\unskip}     \fi
\ifx \showLCCN     \undefined \def \showLCCN      #1{\unskip}     \fi
\ifx \shownote     \undefined \def \shownote      #1{#1}          \fi
\ifx \showarticletitle \undefined \def \showarticletitle #1{#1}   \fi
\ifx \showURL      \undefined \def \showURL       {\relax}        \fi
\providecommand\bibfield[2]{#2}
\providecommand\bibinfo[2]{#2}
\providecommand\natexlab[1]{#1}
\providecommand\showeprint[2][]{arXiv:#2}

\bibitem[Caillot et~al\mbox{.}(2022)]%
        {9732063}
\bibfield{author}{\bibinfo{person}{Antoine Caillot}, \bibinfo{person}{Safa
  Ouerghi}, \bibinfo{person}{Pascal Vasseur}, \bibinfo{person}{Rémi Boutteau},
  {and} \bibinfo{person}{Yohan Dupuis}.} \bibinfo{year}{2022}\natexlab{}.
\newblock \showarticletitle{Survey on Cooperative Perception in an Automotive
  Context}.
\newblock \bibinfo{journal}{\emph{IEEE Transactions on Intelligent
  Transportation Systems}} \bibinfo{volume}{23}, \bibinfo{number}{9}
  (\bibinfo{year}{2022}), \bibinfo{pages}{14204--14223}.
\newblock
\href{https://doi.org/10.1109/TITS.2022.3153815}{doi:\nolinkurl{10.1109/TITS.2022.3153815}}


\bibitem[Chen et~al\mbox{.}(2026)]%
        {cora}
\bibfield{author}{\bibinfo{person}{Gong Chen}, \bibinfo{person}{Chaokun Zhang},
  \bibinfo{person}{Pengcheng Lv}, {and} \bibinfo{person}{Xiaohui Xie}.}
  \bibinfo{year}{2026}\natexlab{}.
\newblock \showarticletitle{CoRA: A Collaborative Robust Architecture with
  Hybrid Fusion for Efficient Perception}.
\newblock \bibinfo{journal}{\emph{Proceedings of the AAAI Conference on
  Artificial Intelligence}} \bibinfo{volume}{40}, \bibinfo{number}{4}
  (\bibinfo{date}{March} \bibinfo{year}{2026}), \bibinfo{pages}{2841--2849}.
\newblock
\href{https://doi.org/10.1609/aaai.v40i4.37274}{doi:\nolinkurl{10.1609/aaai.v40i4.37274}}


\bibitem[Chen et~al\mbox{.}(2019)]%
        {cooper}
\bibfield{author}{\bibinfo{person}{Qi Chen}, \bibinfo{person}{Sihai Tang},
  \bibinfo{person}{Qing Yang}, {and} \bibinfo{person}{Song Fu}.}
  \bibinfo{year}{2019}\natexlab{}.
\newblock \showarticletitle{Cooper: Cooperative Perception for Connected
  Autonomous Vehicles based on 3D Point Clouds}. In
  \bibinfo{booktitle}{\emph{2019 IEEE 39th International Conference on
  Distributed Computing Systems (ICDCS)}} (Dallas, TX, USA).
  \bibinfo{publisher}{IEEE}, \bibinfo{address}{Piscataway, NJ, USA},
  \bibinfo{pages}{514--524}.
\newblock
\href{https://doi.org/10.1109/ICDCS.2019.00058}{doi:\nolinkurl{10.1109/ICDCS.2019.00058}}


\bibitem[Gao et~al\mbox{.}(2024)]%
        {10517450}
\bibfield{author}{\bibinfo{person}{Xin Gao}, \bibinfo{person}{Xinyu Zhang},
  \bibinfo{person}{Yiguo Lu}, \bibinfo{person}{Yuning Huang},
  \bibinfo{person}{Lei Yang}, \bibinfo{person}{Yijin Xiong}, {and}
  \bibinfo{person}{Peng Liu}.} \bibinfo{year}{2024}\natexlab{}.
\newblock \showarticletitle{A Survey of Collaborative Perception in Intelligent
  Vehicles at Intersections}.
\newblock \bibinfo{journal}{\emph{IEEE Transactions on Intelligent Vehicles}}
  (\bibinfo{year}{2024}), \bibinfo{pages}{1--20}.
\newblock
\href{https://doi.org/10.1109/TIV.2024.3395783}{doi:\nolinkurl{10.1109/TIV.2024.3395783}}


\bibitem[Gu et~al\mbox{.}(2023)]%
        {feoco}
\bibfield{author}{\bibinfo{person}{Jiaming Gu}, \bibinfo{person}{Jingyu Zhang},
  \bibinfo{person}{Muyang Zhang}, \bibinfo{person}{Weiliang Meng},
  \bibinfo{person}{Shibiao Xu}, \bibinfo{person}{Jiguang Zhang}, {and}
  \bibinfo{person}{Xiaopeng Zhang}.} \bibinfo{year}{2023}\natexlab{}.
\newblock \showarticletitle{FeaCo: Reaching Robust Feature-Level Consensus in
  Noisy Pose Conditions}. In \bibinfo{booktitle}{\emph{Proceedings of the 31st
  ACM International Conference on Multimedia}} (Ottawa, ON, Canada)
  \emph{(\bibinfo{series}{MM '23})}. \bibinfo{publisher}{Association for
  Computing Machinery}, \bibinfo{address}{New York, NY, USA},
  \bibinfo{pages}{3628--3636}.
\newblock
\showISBNx{9798400701085}
\href{https://doi.org/10.1145/3581783.3611880}{doi:\nolinkurl{10.1145/3581783.3611880}}


\bibitem[Hu et~al\mbox{.}(2022)]%
        {where2comm}
\bibfield{author}{\bibinfo{person}{Yue Hu}, \bibinfo{person}{Shaoheng Fang},
  \bibinfo{person}{Zixing Lei}, \bibinfo{person}{Yiqi Zhong}, {and}
  \bibinfo{person}{Siheng Chen}.} \bibinfo{year}{2022}\natexlab{}.
\newblock \showarticletitle{Where2comm: Communication-Efficient Collaborative
  Perception via Spatial Confidence Maps}. In
  \bibinfo{booktitle}{\emph{Advances in Neural Information Processing Systems}}
  (New Orleans, LA, USA), Vol.~\bibinfo{volume}{35}. \bibinfo{publisher}{Curran
  Associates, Inc.}, \bibinfo{address}{Red Hook, NY, USA},
  \bibinfo{pages}{4874--4886}.
\newblock
\href{https://doi.org/10.52202/068431-0352}{doi:\nolinkurl{10.52202/068431-0352}}


\bibitem[Huang et~al\mbox{.}(2025a)]%
        {survey}
\bibfield{author}{\bibinfo{person}{Tao Huang}, \bibinfo{person}{Jianan Liu},
  \bibinfo{person}{Xi Zhou}, \bibinfo{person}{Dinh~C. Nguyen},
  \bibinfo{person}{Mostafa Rahimi~Azghadi}, \bibinfo{person}{Yuxuan Xia},
  \bibinfo{person}{Qing-Long Han}, {and} \bibinfo{person}{Sumei Sun}.}
  \bibinfo{year}{2025}\natexlab{a}.
\newblock \showarticletitle{Vehicle-to-Everything Cooperative Perception for
  Autonomous Driving}.
\newblock \bibinfo{journal}{\emph{Proc. IEEE}} \bibinfo{volume}{113},
  \bibinfo{number}{5} (\bibinfo{date}{May} \bibinfo{year}{2025}),
  \bibinfo{pages}{443--477}.
\newblock
\showISSN{1558-2256}
\href{https://doi.org/10.1109/JPROC.2025.3600903}{doi:\nolinkurl{10.1109/JPROC.2025.3600903}}


\bibitem[Huang et~al\mbox{.}(2024)]%
        {roco}
\bibfield{author}{\bibinfo{person}{Zhe Huang}, \bibinfo{person}{Shuo Wang},
  \bibinfo{person}{Yongcai Wang}, \bibinfo{person}{Wanting Li},
  \bibinfo{person}{Deying Li}, {and} \bibinfo{person}{Lei Wang}.}
  \bibinfo{year}{2024}\natexlab{}.
\newblock \showarticletitle{RoCo: Robust Cooperative Perception By Iterative
  Object Matching and Pose Adjustment}. In
  \bibinfo{booktitle}{\emph{Proceedings of the 32nd ACM International
  Conference on Multimedia}} (Melbourne, VIC, Australia)
  \emph{(\bibinfo{series}{MM '24})}. \bibinfo{publisher}{Association for
  Computing Machinery}, \bibinfo{address}{New York, NY, USA},
  \bibinfo{pages}{7833--7842}.
\newblock
\href{https://doi.org/10.1145/3664647.3680559}{doi:\nolinkurl{10.1145/3664647.3680559}}


\bibitem[Huang et~al\mbox{.}(2025b)]%
        {codiff}
\bibfield{author}{\bibinfo{person}{Zhe Huang}, \bibinfo{person}{Shuo Wang},
  \bibinfo{person}{Yongcai Wang}, {and} \bibinfo{person}{Lei Wang}.}
  \bibinfo{year}{2025}\natexlab{b}.
\newblock \bibinfo{title}{{CoDiff}: Conditional Diffusion Model for
  Collaborative 3D Object Detection}.
\newblock
\showeprint[arxiv]{2502.14891}~[cs.CV]
\urldef\tempurl%
\url{https://arxiv.org/abs/2502.14891}
\showURL{%
\tempurl}


\bibitem[Kalman(1960)]%
        {kalman}
\bibfield{author}{\bibinfo{person}{Rudolph~Emil Kalman}.}
  \bibinfo{year}{1960}\natexlab{}.
\newblock \showarticletitle{A New Approach to Linear Filtering and Prediction
  Problems}.
\newblock \bibinfo{journal}{\emph{Transactions of the ASME--Journal of Basic
  Engineering}} \bibinfo{volume}{82}, \bibinfo{number}{1}
  (\bibinfo{year}{1960}), \bibinfo{pages}{35--45}.
\newblock
\href{https://doi.org/10.1115/1.3662552}{doi:\nolinkurl{10.1115/1.3662552}}


\bibitem[Kuhn(1955)]%
        {hungarian}
\bibfield{author}{\bibinfo{person}{H.~W. Kuhn}.}
  \bibinfo{year}{1955}\natexlab{}.
\newblock \showarticletitle{The Hungarian method for the assignment problem}.
\newblock \bibinfo{journal}{\emph{Naval Research Logistics Quarterly}}
  \bibinfo{volume}{2}, \bibinfo{number}{1-2} (\bibinfo{year}{1955}),
  \bibinfo{pages}{83--97}.
\newblock
\href{https://doi.org/10.1002/nav.3800020109}{doi:\nolinkurl{10.1002/nav.3800020109}}


\bibitem[Lang et~al\mbox{.}(2019)]%
        {pointpillars}
\bibfield{author}{\bibinfo{person}{Alex~H. Lang}, \bibinfo{person}{Sourabh
  Vora}, \bibinfo{person}{Holger Caesar}, \bibinfo{person}{Lubing Zhou},
  \bibinfo{person}{Jiong Yang}, {and} \bibinfo{person}{Oscar Beijbom}.}
  \bibinfo{year}{2019}\natexlab{}.
\newblock \showarticletitle{PointPillars: Fast Encoders for Object Detection
  from Point Clouds}. In \bibinfo{booktitle}{\emph{Proceedings of the IEEE/CVF
  Conference on Computer Vision and Pattern Recognition (CVPR)}} (Long Beach,
  CA, USA). \bibinfo{publisher}{IEEE}, \bibinfo{address}{Piscataway, NJ, USA},
  \bibinfo{pages}{12697--12705}.
\newblock
\href{https://doi.org/10.1109/CVPR.2019.01298}{doi:\nolinkurl{10.1109/CVPR.2019.01298}}


\bibitem[Lei et~al\mbox{.}(2024)]%
        {freeAlign}
\bibfield{author}{\bibinfo{person}{Zixing Lei}, \bibinfo{person}{Zhenyang Ni},
  \bibinfo{person}{Ruize Han}, \bibinfo{person}{Shuo Tang},
  \bibinfo{person}{Dingju Wang}, \bibinfo{person}{Chen Feng},
  \bibinfo{person}{Siheng Chen}, {and} \bibinfo{person}{Yanfeng Wang}.}
  \bibinfo{year}{2024}\natexlab{}.
\newblock \showarticletitle{Robust Collaborative Perception without External
  Localization and Clock Devices}. In \bibinfo{booktitle}{\emph{2024 IEEE
  International Conference on Robotics and Automation (ICRA)}} (Yokohama,
  Japan). \bibinfo{publisher}{IEEE}, \bibinfo{address}{Piscataway, NJ, USA},
  \bibinfo{pages}{7280--7286}.
\newblock
\showISBNx{979-8-3503-8457-4}
\href{https://doi.org/10.1109/ICRA57147.2024.10610635}{doi:\nolinkurl{10.1109/ICRA57147.2024.10610635}}


\bibitem[Lei et~al\mbox{.}(2022)]%
        {syncnet}
\bibfield{author}{\bibinfo{person}{Zixing Lei}, \bibinfo{person}{Shunli Ren},
  \bibinfo{person}{Yue Hu}, \bibinfo{person}{Wenjun Zhang}, {and}
  \bibinfo{person}{Siheng Chen}.} \bibinfo{year}{2022}\natexlab{}.
\newblock \showarticletitle{Latency-Aware Collaborative Perception}. In
  \bibinfo{booktitle}{\emph{Computer Vision -- ECCV 2022: 17th European
  Conference, Tel Aviv, Israel, October 23--27, 2022, Proceedings, Part XXXII}}
  (Tel Aviv, Israel) \emph{(\bibinfo{series}{Lecture Notes in Computer
  Science}, Vol.~\bibinfo{volume}{13692})}. \bibinfo{publisher}{Springer},
  \bibinfo{address}{Cham, Switzerland}, \bibinfo{pages}{316--332}.
\newblock
\showISBNx{978-3-031-19823-6}
\href{https://doi.org/10.1007/978-3-031-19824-3_19}{doi:\nolinkurl{10.1007/978-3-031-19824-3_19}}


\bibitem[Lu et~al\mbox{.}(2023)]%
        {coalign}
\bibfield{author}{\bibinfo{person}{Yifan Lu}, \bibinfo{person}{Quanhao Li},
  \bibinfo{person}{Baoan Liu}, \bibinfo{person}{Mehrdad Dianati},
  \bibinfo{person}{Chen Feng}, \bibinfo{person}{Siheng Chen}, {and}
  \bibinfo{person}{Yanfeng Wang}.} \bibinfo{year}{2023}\natexlab{}.
\newblock \showarticletitle{Robust collaborative {{}{3D}} object detection in
  presence of pose errors}. In \bibinfo{booktitle}{\emph{2023 IEEE
  International Conference on Robotics and Automation (ICRA)}} (London, United
  Kingdom). \bibinfo{publisher}{IEEE}, \bibinfo{address}{Piscataway, NJ, USA},
  \bibinfo{pages}{4812--4818}.
\newblock
\showISBNx{979-8-3503-2366-5}
\href{https://doi.org/10.1109/ICRA48891.2023.10160546}{doi:\nolinkurl{10.1109/ICRA48891.2023.10160546}}


\bibitem[Ni et~al\mbox{.}(2024)]%
        {cobevglue}
\bibfield{author}{\bibinfo{person}{Zhenyang Ni}, \bibinfo{person}{Zixing Lei},
  \bibinfo{person}{Yifan Lu}, \bibinfo{person}{Dingju Wang},
  \bibinfo{person}{Chen Feng}, \bibinfo{person}{Yanfeng Wang}, {and}
  \bibinfo{person}{Siheng Chen}.} \bibinfo{year}{2024}\natexlab{}.
\newblock \bibinfo{title}{Self-Localized Collaborative Perception}.
\newblock
\showeprint[arxiv]{2406.12712}~[cs.CV]
\href{https://doi.org/10.48550/arXiv.2406.12712}{doi:\nolinkurl{10.48550/arXiv.2406.12712}}


\bibitem[Rauch et~al\mbox{.}(2012)]%
        {6232130}
\bibfield{author}{\bibinfo{person}{Andreas Rauch}, \bibinfo{person}{Felix
  Klanner}, \bibinfo{person}{Ralph Rasshofer}, {and} \bibinfo{person}{Klaus
  Dietmayer}.} \bibinfo{year}{2012}\natexlab{}.
\newblock \showarticletitle{Car2X-based perception in a high-level fusion
  architecture for cooperative perception systems}. In
  \bibinfo{booktitle}{\emph{2012 IEEE Intelligent Vehicles Symposium}}
  (Alcal{\'a} de Henares, Spain). \bibinfo{publisher}{IEEE},
  \bibinfo{address}{Piscataway, NJ, USA}, \bibinfo{pages}{270--275}.
\newblock
\href{https://doi.org/10.1109/IVS.2012.6232130}{doi:\nolinkurl{10.1109/IVS.2012.6232130}}


\bibitem[Rawashdeh and Wang(2018)]%
        {8569832}
\bibfield{author}{\bibinfo{person}{Zaydoun~Yahya Rawashdeh} {and}
  \bibinfo{person}{Zheng Wang}.} \bibinfo{year}{2018}\natexlab{}.
\newblock \showarticletitle{Collaborative Automated Driving: A Machine
  Learning-based Method to Enhance the Accuracy of Shared Information}. In
  \bibinfo{booktitle}{\emph{2018 21st International Conference on Intelligent
  Transportation Systems (ITSC)}} (Maui, HI, USA). \bibinfo{publisher}{IEEE},
  \bibinfo{address}{Piscataway, NJ, USA}, \bibinfo{pages}{3961--3966}.
\newblock
\href{https://doi.org/10.1109/ITSC.2018.8569832}{doi:\nolinkurl{10.1109/ITSC.2018.8569832}}


\bibitem[Song et~al\mbox{.}(2025)]%
        {traf_align}
\bibfield{author}{\bibinfo{person}{Zhiying Song}, \bibinfo{person}{Lei Yang},
  \bibinfo{person}{Fuxi Wen}, {and} \bibinfo{person}{Jun Li}.}
  \bibinfo{year}{2025}\natexlab{}.
\newblock \showarticletitle{{{}{TraF-Align}}: Trajectory-aware feature
  alignment for asynchronous multi-agent perception}. In
  \bibinfo{booktitle}{\emph{Proceedings of the IEEE/CVF Conference on Computer
  Vision and Pattern Recognition (CVPR)}} (Nashville, TN, USA).
  \bibinfo{publisher}{IEEE}, \bibinfo{address}{Piscataway, NJ, USA},
  \bibinfo{pages}{12048--12057}.
\newblock
\href{https://doi.org/10.1109/CVPR52734.2025.01125}{doi:\nolinkurl{10.1109/CVPR52734.2025.01125}}


\bibitem[Tang et~al\mbox{.}(2025)]%
        {cost}
\bibfield{author}{\bibinfo{person}{Zongheng Tang}, \bibinfo{person}{Yi Liu},
  \bibinfo{person}{Yifan Sun}, \bibinfo{person}{Yulu Gao},
  \bibinfo{person}{Jinyu Chen}, \bibinfo{person}{Runsheng Xu}, {and}
  \bibinfo{person}{Si Liu}.} \bibinfo{year}{2025}\natexlab{}.
\newblock \showarticletitle{CoST: Efficient Collaborative Perception From
  Unified Spatiotemporal Perspective}. In \bibinfo{booktitle}{\emph{Proceedings
  of the IEEE/CVF International Conference on Computer Vision (ICCV)}}
  (Honolulu, HI, USA). \bibinfo{publisher}{IEEE}, \bibinfo{address}{Piscataway,
  NJ, USA}, \bibinfo{pages}{1120--1129}.
\newblock
\urldef\tempurl%
\url{https://openaccess.thecvf.com/content/ICCV2025/html/Tang_CoST_Efficient_Collaborative_Perception_From_Unified_Spatiotemporal_Perspective_ICCV_2025_paper.html}
\showURL{%
\tempurl}


\bibitem[Vadivelu et~al\mbox{.}(2021)]%
        {v2vnetro}
\bibfield{author}{\bibinfo{person}{Nicholas Vadivelu}, \bibinfo{person}{Mengye
  Ren}, \bibinfo{person}{James Tu}, \bibinfo{person}{Jingkang Wang}, {and}
  \bibinfo{person}{Raquel Urtasun}.} \bibinfo{year}{2021}\natexlab{}.
\newblock \showarticletitle{Learning to Communicate and Correct Pose Errors}.
  In \bibinfo{booktitle}{\emph{Proceedings of the 2020 Conference on Robot
  Learning}} (Virtual Conference) \emph{(\bibinfo{series}{Proceedings of
  Machine Learning Research}, Vol.~\bibinfo{volume}{155})},
  \bibfield{editor}{\bibinfo{person}{Jens Kober}, \bibinfo{person}{Fabio
  Ramos}, {and} \bibinfo{person}{Claire Tomlin}} (Eds.).
  \bibinfo{publisher}{PMLR}, \bibinfo{address}{Cambridge, MA, USA},
  \bibinfo{pages}{1195--1210}.
\newblock
\urldef\tempurl%
\url{https://proceedings.mlr.press/v155/vadivelu21a.html}
\showURL{%
\tempurl}


\bibitem[Wan et~al\mbox{.}(2026)]%
        {survey1}
\bibfield{author}{\bibinfo{person}{Lei Wan}, \bibinfo{person}{Jianxin Zhao},
  \bibinfo{person}{Andreas Wiedholz}, \bibinfo{person}{Manuel Bied},
  \bibinfo{person}{Mateus Martinez~de Lucena}, \bibinfo{person}{Abhishek
  Dinkar~Jagtap}, \bibinfo{person}{Andreas Festag}, \bibinfo{person}{Antônio
  Augusto~Fröhlich}, \bibinfo{person}{Hannan Ejaz~Keen}, {and}
  \bibinfo{person}{Alexey Vinel}.} \bibinfo{year}{2026}\natexlab{}.
\newblock \showarticletitle{A Systematic Literature Review on Vehicular
  Collaborative Perception—A Computer Vision Perspective}.
\newblock \bibinfo{journal}{\emph{IEEE Transactions on Intelligent
  Transportation Systems}} \bibinfo{volume}{27}, \bibinfo{number}{1}
  (\bibinfo{date}{Jan.} \bibinfo{year}{2026}), \bibinfo{pages}{81--118}.
\newblock
\showISSN{1558-0016}
\href{https://doi.org/10.1109/TITS.2025.3631141}{doi:\nolinkurl{10.1109/TITS.2025.3631141}}


\bibitem[Wang and Nordström(2025)]%
        {lrcp}
\bibfield{author}{\bibinfo{person}{Junjie Wang} {and} \bibinfo{person}{Tomas
  Nordström}.} \bibinfo{year}{2025}\natexlab{}.
\newblock \showarticletitle{Latency Robust Cooperative Perception Using
  Asynchronous Feature Fusion}. In \bibinfo{booktitle}{\emph{2025 IEEE/CVF
  Winter Conference on Applications of Computer Vision (WACV)}} (Tucson, AZ,
  USA). \bibinfo{publisher}{IEEE}, \bibinfo{address}{Piscataway, NJ, USA},
  \bibinfo{pages}{4862--4871}.
\newblock
\href{https://doi.org/10.1109/WACV61041.2025.00476}{doi:\nolinkurl{10.1109/WACV61041.2025.00476}}


\bibitem[Wang et~al\mbox{.}(2025)]%
        {v2xdgpe}
\bibfield{author}{\bibinfo{person}{Sichao Wang}, \bibinfo{person}{Ming Yuan},
  \bibinfo{person}{Chuang Zhang}, \bibinfo{person}{Qing Xu},
  \bibinfo{person}{Lei He}, {and} \bibinfo{person}{Jianqiang Wang}.}
  \bibinfo{year}{2025}\natexlab{}.
\newblock \showarticletitle{V2X-DGPE: Addressing Domain Gaps and Pose Errors
  for Robust Collaborative 3D Object Detection}. In
  \bibinfo{booktitle}{\emph{2025 IEEE Intelligent Vehicles Symposium (IV)}}
  (Cluj-Napoca, Romania). \bibinfo{publisher}{IEEE},
  \bibinfo{address}{Piscataway, NJ, USA}, \bibinfo{pages}{2074--2080}.
\newblock
\href{https://doi.org/10.1109/IV64158.2025.11097385}{doi:\nolinkurl{10.1109/IV64158.2025.11097385}}


\bibitem[Wei et~al\mbox{.}(2023)]%
        {cobevflow}
\bibfield{author}{\bibinfo{person}{Sizhe Wei}, \bibinfo{person}{Yuxi Wei},
  \bibinfo{person}{Yue Hu}, \bibinfo{person}{Yifan Lu}, \bibinfo{person}{Yiqi
  Zhong}, \bibinfo{person}{Siheng Chen}, {and} \bibinfo{person}{Ya Zhang}.}
  \bibinfo{year}{2023}\natexlab{}.
\newblock \showarticletitle{Asynchrony-Robust Collaborative Perception via
  Bird's Eye View Flow}. In \bibinfo{booktitle}{\emph{Advances in Neural
  Information Processing Systems}} (New Orleans, LA, USA),
  Vol.~\bibinfo{volume}{36}. \bibinfo{publisher}{Curran Associates, Inc.},
  \bibinfo{address}{Red Hook, NY, USA}, \bibinfo{pages}{28462--28477}.
\newblock
\href{https://doi.org/10.52202/075280-1236}{doi:\nolinkurl{10.52202/075280-1236}}


\bibitem[Xu et~al\mbox{.}(2025b)]%
        {sonata}
\bibfield{author}{\bibinfo{person}{Dongzhu Xu}, \bibinfo{person}{Rui Lin},
  \bibinfo{person}{Huanhuan Zhang}, \bibinfo{person}{Anfu Zhou}, {and}
  \bibinfo{person}{Huadong Ma}.} \bibinfo{year}{2025}\natexlab{b}.
\newblock \showarticletitle{Bridging Cross-Layer Interactions Between 5G RAN
  and MEC for Latency-Critical Video Analytics}.
\newblock \bibinfo{journal}{\emph{IEEE Transactions on Networking}}
  \bibinfo{volume}{33}, \bibinfo{number}{4} (\bibinfo{year}{2025}),
  \bibinfo{pages}{1614--1629}.
\newblock
\href{https://doi.org/10.1109/TON.2025.3542456}{doi:\nolinkurl{10.1109/TON.2025.3542456}}


\bibitem[Xu et~al\mbox{.}(2023a)]%
        {cobevt}
\bibfield{author}{\bibinfo{person}{Runsheng Xu}, \bibinfo{person}{Zhengzhong
  Tu}, \bibinfo{person}{Hao Xiang}, \bibinfo{person}{Wei Shao},
  \bibinfo{person}{Bolei Zhou}, {and} \bibinfo{person}{Jiaqi Ma}.}
  \bibinfo{year}{2023}\natexlab{a}.
\newblock \showarticletitle{CoBEVT: Cooperative Bird's Eye View Semantic
  Segmentation with Sparse Transformers}. In
  \bibinfo{booktitle}{\emph{Proceedings of the 6th Conference on Robot
  Learning}} (Auckland, New Zealand) \emph{(\bibinfo{series}{Proceedings of
  Machine Learning Research}, Vol.~\bibinfo{volume}{205})},
  \bibfield{editor}{\bibinfo{person}{Karen Liu}, \bibinfo{person}{Dana Kulic},
  {and} \bibinfo{person}{Jeff Ichnowski}} (Eds.). \bibinfo{publisher}{PMLR},
  \bibinfo{address}{Cambridge, MA, USA}, \bibinfo{pages}{989--1000}.
\newblock
\urldef\tempurl%
\url{https://proceedings.mlr.press/v205/xu23a.html}
\showURL{%
\tempurl}


\bibitem[Xu et~al\mbox{.}(2023b)]%
        {v2v4real}
\bibfield{author}{\bibinfo{person}{Runsheng Xu}, \bibinfo{person}{Xin Xia},
  \bibinfo{person}{Jinlong Li}, \bibinfo{person}{Hanzhao Li},
  \bibinfo{person}{Shuo Zhang}, \bibinfo{person}{Zhengzhong Tu},
  \bibinfo{person}{Zonglin Meng}, \bibinfo{person}{Hao Xiang},
  \bibinfo{person}{Xiaoyu Dong}, \bibinfo{person}{Rui Song},
  \bibinfo{person}{Hongkai Yu}, \bibinfo{person}{Bolei Zhou}, {and}
  \bibinfo{person}{Jiaqi Ma}.} \bibinfo{year}{2023}\natexlab{b}.
\newblock \showarticletitle{V2V4Real: A Real-world Large-scale Dataset for
  Vehicle-to-Vehicle Cooperative Perception}. In
  \bibinfo{booktitle}{\emph{Proceedings of the IEEE/CVF Conference on Computer
  Vision and Pattern Recognition (CVPR)}} (Vancouver, BC, Canada).
  \bibinfo{publisher}{IEEE}, \bibinfo{address}{Piscataway, NJ, USA},
  \bibinfo{pages}{13712--13722}.
\newblock
\href{https://doi.org/10.1109/CVPR52729.2023.01318}{doi:\nolinkurl{10.1109/CVPR52729.2023.01318}}


\bibitem[Xu et~al\mbox{.}(2022a)]%
        {v2xvit}
\bibfield{author}{\bibinfo{person}{Runsheng Xu}, \bibinfo{person}{Hao Xiang},
  \bibinfo{person}{Zhengzhong Tu}, \bibinfo{person}{Xin Xia},
  \bibinfo{person}{Ming-Hsuan Yang}, {and} \bibinfo{person}{Jiaqi Ma}.}
  \bibinfo{year}{2022}\natexlab{a}.
\newblock \showarticletitle{V2X-ViT: Vehicle-to-Everything Cooperative
  Perception with Vision Transformer}. In \bibinfo{booktitle}{\emph{Computer
  Vision -- ECCV 2022: 17th European Conference, Tel Aviv, Israel, October
  23--27, 2022, Proceedings, Part XXXIX}} (Tel Aviv, Israel)
  \emph{(\bibinfo{series}{Lecture Notes in Computer Science},
  Vol.~\bibinfo{volume}{13699})}. \bibinfo{publisher}{Springer},
  \bibinfo{address}{Cham, Switzerland}, \bibinfo{pages}{107--124}.
\newblock
\href{https://doi.org/10.1007/978-3-031-19842-7_7}{doi:\nolinkurl{10.1007/978-3-031-19842-7_7}}


\bibitem[Xu et~al\mbox{.}(2022b)]%
        {opv2v}
\bibfield{author}{\bibinfo{person}{Runsheng Xu}, \bibinfo{person}{Hao Xiang},
  \bibinfo{person}{Xin Xia}, \bibinfo{person}{Xu Han}, \bibinfo{person}{Jinlong
  Li}, {and} \bibinfo{person}{Jiaqi Ma}.} \bibinfo{year}{2022}\natexlab{b}.
\newblock \showarticletitle{OPV2V: An Open Benchmark Dataset and Fusion
  Pipeline for Perception with Vehicle-to-Vehicle Communication}. In
  \bibinfo{booktitle}{\emph{2022 International Conference on Robotics and
  Automation (ICRA)}} (Philadelphia, PA, USA). \bibinfo{publisher}{IEEE},
  \bibinfo{address}{Piscataway, NJ, USA}, \bibinfo{pages}{2583--2589}.
\newblock
\showISBNx{978-1-7281-9681-7}
\href{https://doi.org/10.1109/ICRA46639.2022.9812038}{doi:\nolinkurl{10.1109/ICRA46639.2022.9812038}}


\bibitem[Xu et~al\mbox{.}(2025a)]%
        {codyntrust}
\bibfield{author}{\bibinfo{person}{Yunjiang Xu}, \bibinfo{person}{Lingzhi Li},
  \bibinfo{person}{Jin Wang}, \bibinfo{person}{Benyuan Yang},
  \bibinfo{person}{Zhiwen Wu}, \bibinfo{person}{Xinhong Chen}, {and}
  \bibinfo{person}{Jianping Wang}.} \bibinfo{year}{2025}\natexlab{a}.
\newblock \showarticletitle{CoDynTrust: Robust Asynchronous Collaborative
  Perception via Dynamic Feature Trust Modulus}. In
  \bibinfo{booktitle}{\emph{2025 IEEE International Conference on Robotics and
  Automation (ICRA)}} (Atlanta, GA, USA). \bibinfo{publisher}{IEEE},
  \bibinfo{address}{Piscataway, NJ, USA}, \bibinfo{pages}{336--342}.
\newblock
\href{https://doi.org/10.1109/ICRA55743.2025.11127779}{doi:\nolinkurl{10.1109/ICRA55743.2025.11127779}}


\bibitem[Yang et~al\mbox{.}(2023)]%
        {scope}
\bibfield{author}{\bibinfo{person}{Kun Yang}, \bibinfo{person}{Dingkang Yang},
  \bibinfo{person}{Jingyu Zhang}, \bibinfo{person}{Mingcheng Li},
  \bibinfo{person}{Yang Liu}, \bibinfo{person}{Jing Liu},
  \bibinfo{person}{Hanqi Wang}, \bibinfo{person}{Peng Sun}, {and}
  \bibinfo{person}{Liang Song}.} \bibinfo{year}{2023}\natexlab{}.
\newblock \showarticletitle{Spatio-Temporal Domain Awareness for Multi-Agent
  Collaborative Perception}. In \bibinfo{booktitle}{\emph{Proceedings of the
  IEEE/CVF International Conference on Computer Vision (ICCV)}} (Paris,
  France). \bibinfo{publisher}{IEEE}, \bibinfo{address}{Piscataway, NJ, USA},
  \bibinfo{pages}{23383--23392}.
\newblock
\href{https://doi.org/10.1109/ICCV51070.2023.02137}{doi:\nolinkurl{10.1109/ICCV51070.2023.02137}}


\bibitem[Yu et~al\mbox{.}(2023)]%
        {ffnet}
\bibfield{author}{\bibinfo{person}{Haibao Yu}, \bibinfo{person}{Yingjuan Tang},
  \bibinfo{person}{Enze Xie}, \bibinfo{person}{Jilei Mao},
  \bibinfo{person}{Ping Luo}, {and} \bibinfo{person}{Zaiqing Nie}.}
  \bibinfo{year}{2023}\natexlab{}.
\newblock \showarticletitle{Flow-based feature fusion for
  vehicle-infrastructure cooperative 3D object detection}. In
  \bibinfo{booktitle}{\emph{Advances in Neural Information Processing Systems}}
  (New Orleans, LA, USA), Vol.~\bibinfo{volume}{36}. \bibinfo{publisher}{Curran
  Associates, Inc.}, \bibinfo{address}{Red Hook, NY, USA},
  \bibinfo{pages}{34493--34503}.
\newblock
\href{https://doi.org/10.52202/075280-1497}{doi:\nolinkurl{10.52202/075280-1497}}


\bibitem[Yuan et~al\mbox{.}(2022)]%
        {fpv-rcnn}
\bibfield{author}{\bibinfo{person}{Yunshuang Yuan}, \bibinfo{person}{Hao
  Cheng}, {and} \bibinfo{person}{Monika Sester}.}
  \bibinfo{year}{2022}\natexlab{}.
\newblock \showarticletitle{Keypoints-Based Deep Feature Fusion for Cooperative
  Vehicle Detection of Autonomous Driving}.
\newblock \bibinfo{journal}{\emph{IEEE Robotics and Automation Letters}}
  \bibinfo{volume}{7}, \bibinfo{number}{2} (\bibinfo{year}{2022}),
  \bibinfo{pages}{3054--3061}.
\newblock
\href{https://doi.org/10.1109/LRA.2022.3143299}{doi:\nolinkurl{10.1109/LRA.2022.3143299}}


\bibitem[Yuan et~al\mbox{.}(2025)]%
        {sparsealign}
\bibfield{author}{\bibinfo{person}{Yunshuang Yuan}, \bibinfo{person}{Yan Xia},
  \bibinfo{person}{Daniel Cremers}, {and} \bibinfo{person}{Monika Sester}.}
  \bibinfo{year}{2025}\natexlab{}.
\newblock \showarticletitle{SparseAlign: A Fully Sparse Framework for
  Cooperative Object Detection}. In \bibinfo{booktitle}{\emph{Proceedings of
  the IEEE/CVF Conference on Computer Vision and Pattern Recognition (CVPR)}}
  (Nashville, TN, USA). \bibinfo{publisher}{IEEE},
  \bibinfo{address}{Piscataway, NJ, USA}, \bibinfo{pages}{22296--22305}.
\newblock
\href{https://doi.org/10.1109/CVPR52734.2025.02077}{doi:\nolinkurl{10.1109/CVPR52734.2025.02077}}


\bibitem[Zhang et~al\mbox{.}(2024)]%
        {ermvp}
\bibfield{author}{\bibinfo{person}{Jingyu Zhang}, \bibinfo{person}{Kun Yang},
  \bibinfo{person}{Yilei Wang}, \bibinfo{person}{Hanqi Wang},
  \bibinfo{person}{Peng Sun}, {and} \bibinfo{person}{Liang Song}.}
  \bibinfo{year}{2024}\natexlab{}.
\newblock \showarticletitle{ERMVP: Communication-Efficient and
  Collaboration-Robust Multi-Vehicle Perception in Challenging Environments}.
  In \bibinfo{booktitle}{\emph{2024 IEEE/CVF Conference on Computer Vision and
  Pattern Recognition (CVPR)}} (Seattle, WA, USA). \bibinfo{publisher}{IEEE},
  \bibinfo{address}{Piscataway, NJ, USA}, \bibinfo{pages}{12575--12584}.
\newblock
\href{https://doi.org/10.1109/CVPR52733.2024.01195}{doi:\nolinkurl{10.1109/CVPR52733.2024.01195}}


\bibitem[Zhou et~al\mbox{.}(2025)]%
        {v2xpnp}
\bibfield{author}{\bibinfo{person}{Zewei Zhou}, \bibinfo{person}{Hao Xiang},
  \bibinfo{person}{Zhaoliang Zheng}, \bibinfo{person}{Seth~Z. Zhao},
  \bibinfo{person}{Mingyue Lei}, \bibinfo{person}{Yun Zhang},
  \bibinfo{person}{Tianhui Cai}, \bibinfo{person}{Xinyi Liu},
  \bibinfo{person}{Johnson Liu}, \bibinfo{person}{Maheswari Bajji},
  \bibinfo{person}{Xin Xia}, \bibinfo{person}{Zhiyu Huang},
  \bibinfo{person}{Bolei Zhou}, {and} \bibinfo{person}{Jiaqi Ma}.}
  \bibinfo{year}{2025}\natexlab{}.
\newblock \showarticletitle{V2XPnP: Vehicle-to-Everything Spatio-Temporal
  Fusion for Multi-Agent Perception and Prediction}. In
  \bibinfo{booktitle}{\emph{Proceedings of the IEEE/CVF International
  Conference on Computer Vision (ICCV)}} (Honolulu, HI, USA).
  \bibinfo{publisher}{IEEE}, \bibinfo{address}{Piscataway, NJ, USA},
  \bibinfo{pages}{25399--25409}.
\newblock
\href{https://doi.org/10.1109/ICCV51701.2025.02356}{doi:\nolinkurl{10.1109/ICCV51701.2025.02356}}


\end{thebibliography}

\clearpage
\appendix
\section{Evaluation Protocol and Baseline Construction}
\label{sec:app-evaluation}

\subsection{AP Computation}

We report AP@0.5 and AP@0.7 using dataset-level global sorting.
Concretely, after decoding and non-maximum suppression, all predicted boxes over the entire evaluation split are pooled together and sorted by confidence to form a single precision--recall curve at each IoU threshold. 
In other words, the ranking is established once over the whole benchmark, rather than separately within each scene. 
This evaluation setup is also consistent with prior methods such as CoAlign~\cite{coalign} and RoCo~\cite{roco}.

By contrast, some implementations use scene-wise sorting with scene-level aggregation: predictions are first ranked within each scene, and true/false positives are accumulated locally before the results are aggregated scene by scene. 
Accordingly, the absolute AP values under the two protocols can differ numerically. All results in the main text and this appendix follow the same dataset-level global-sorting protocol.

\subsection{Construction of the Cascaded Baseline \base}

We clarify how the cascaded baseline \base is constructed from CoAlign and TraF-Align. Starting from the original TraF-Align pipeline, we insert a CoAlign-style pose-refinement stage before temporal delay compensation. To preserve the co-visibility assumption underlying box-based cross-agent matching, the relative pose is refined only from the delayed neighbor observation and the ego observation at the same delayed timestamp $u_j = t - \tau_j$. In other words, detections from different timestamps are not jointly fed into the CoAlign solver.

After this delayed-time pose refinement, the refined relative pose is used consistently for the delayed collaborator message and for the historical features required by TraF-Align. Specifically, we transform the entire short history into the ego frame using the same refined delayed-time pose together with ego-motion, rather than optimizing each historical frame independently. This design avoids introducing additional frame-wise temporal inconsistency from the inserted spatial-alignment stage, and keeps the cascaded implementation as faithful as possible to the original temporal-compensation pipeline of TraF-Align.

\section{Implementation and Reproducibility Details}
\label{sec:supp_impl}

This appendix section provides implementation details beyond the main text, including the detector architecture, method-specific training configurations, runtime pipeline of the object-level branch, and the full set of rule-based hyperparameters. We first summarize the detector instantiations and optimization settings used in our experiments. We then describe the inference-time flow of the object-level branch and present the corresponding explicit formulation.

\subsection{Method-specific Training Configurations}

Table~\ref{tab:app_method_training} summarizes the optimization schedules and training-time perturbation settings used by the compared methods. 
All methods use AdamW with weight decay $10^{-4}$, and all experiments share the same standard point-cloud augmentation, including scaling, rotation, and flipping.

Although all methods are implemented within the same PointPillar family, their detector-side configurations and training schedules are not strictly identical.
The single-agent detector adopts a PointPillar-style architecture. Voxelized LiDAR points are first encoded by a 64-channel PillarVFE, scattered to the BEV plane, and processed by a three-stage ResNet-style BEV backbone. The multiscale features are then decoded by an upsampling path, compressed by a 256-channel shrink head, and passed to the standard classification, box-regression, and direction-classification heads. In our implementation, the single-agent detector uses the same backbone widths as the collaborative detector (64--128--256), so that the comparison isolates the effect of collaboration rather than changes in the detector trunk.

The collaborative detector shares the same voxel encoder and BEV backbone, but introduces attention-based intermediate fusion at the three BEV scales. Specifically, per-agent BEV features with channel dimensions 64, 128, and 256 are transformed into the ego frame, fused scale by scale via self-attention, decoded into a shared BEV representation, compressed to 256 channels, and finally fed to the same detection heads. The detector architecture is kept unchanged across datasets.

\begin{table*}[t]
    \centering
    \small
    \setlength{\tabcolsep}{3.5pt}
    \caption{Method-specific training configurations.}
    \label{tab:app_method_training}
    \renewcommand{\arraystretch}{1.06}
    \begin{tabular}{L{2.9cm}|L{2.1cm}|L{5.1cm}|L{5.0cm}}
        \toprule
        \textbf{Method} & \textbf{Epochs} & \textbf{LR / scheduler} & \textbf{Perturbation during training} \\
        \midrule
        \name~(single-agent) & 30 & \makecell[l]{Multi-step decay; initial lr $=0.002$\\$\gamma=0.1$, milestones [10, 20]} & perfect setting throughout \\
        \midrule
        \makecell[l]{\name\\CoAlign} & 30 & \makecell[l]{Multi-step decay; initial lr $=0.002$\\$\gamma=0.1$, milestones [10, 20]} & pose noise $(0.2\,\text{m}, 0.2^\circ)$ \\
        \midrule
        V2X-ViT & \makecell[l]{60 clean\\+ 10 fine-tune} & \makecell[l]{Multi-step decay; initial lr $=0.001$\\$\gamma=0.1$, milestones [15, 50]} & \makecell[l]{Clean pretraining; fine-tune with\\pose noise $(0.2\,\text{m}, 0.2^\circ)$ and\\random delay 200--300\,ms} \\
        \midrule
        \makecell[l]{TraF-Align\\\base} & 60 & \makecell[l]{One-cycle schedule; peak lr $=10^{-4}$\\initial lr $=10^{-5}$; ramp-up fraction $=0.4$} & random delay 0--400\,ms \\
        \midrule
        \makecell[l]{ERMVP\\CoST} & 60 & \makecell[l]{Cosine annealing with warmup\\warmup lr $=2\times10^{-4}$\\warmup epochs $=10$\\minimum lr $=2\times10^{-5}$} & perfect setting throughout \\
        \bottomrule
    \end{tabular}
\end{table*}

For our single-agent detector and our collaborative model, the detection loss follows the standard PointPillar formulation used in the codebase.
Specifically, the positive classification weight is set to $1.0$.
The classification branch uses Sigmoid Focal Loss with $\alpha=0.25$, $\gamma=2.0$, and loss weight $1.0$.
The box-regression branch uses codewise Weighted Smooth L1 Loss with $\sigma=3.0$ and loss weight $2.0$.
The direction branch uses Weighted Softmax Classification Loss with loss weight $0.2$.
The baselines follow the official loss definitions in their released implementations.
\subsection{Formulation in Practice}

The learnable detector/fusion backbone remains unchanged; the proposed object-level module is an inference-time rule-based branch described below.

\textbf{Matching cost.}
Let $\Omega_{ab}$ be the set of common history offsets shared by ego track $a$ and delayed collaborator track $b$, after rotating the collaborator velocity into the ego frame using the initial pose rotation $R_0$. The velocity consistency term is defined as
\begin{equation}
d^{\mathrm{vel}}_{ab}
=
\frac{1}{|\Omega_{ab}|}
\sum_{\kappa\in\Omega_{ab}}
\left\|
v^{e}_{a,\kappa}
-R_0v_{b,\kappa}
\right\|_2 .
\end{equation}
For local-graph consistency, the implementation constructs two signatures from the $K$ nearest neighbors of each track center: an edge-distance signature and a pairwise-neighbor signature. Let $D_{\mathrm{sort}}(\cdot,\cdot)$ denote the mean absolute difference between the sorted values of two signatures truncated to their common length. The graph term is
\begin{equation}
d^{\mathrm{graph}}_{ab}
=
\frac{
w_{\mathrm{edge}}D_{\mathrm{sort}}(s^{\mathrm{edge}}_a,s^{\mathrm{edge}}_b)
+
w_{\mathrm{pair}}D_{\mathrm{sort}}(s^{\mathrm{pair}}_a,s^{\mathrm{pair}}_b)
}{
w_{\mathrm{edge}}+w_{\mathrm{pair}}
}.
\end{equation}
The final matching cost before Hungarian assignment is
\begin{equation}
c_{ab}
=
\lambda_{\mathrm{vel}}\frac{d^{\mathrm{vel}}_{ab}}{T_{\mathrm{vel}}}
+
\lambda_{\mathrm{graph}}\frac{d^{\mathrm{graph}}_{ab}}{T_{\mathrm{graph}}}.
\label{eq:app_match_cost}
\end{equation}

\textbf{Cost-to-pair-weight mapping.}
Pose refinement is executed only when the matched set for neighbor $j$ contains at least $N_{\mathrm{trk}}^{\min}=3$ trajectory correspondences, i.e., at least three matched trajectories are available. For a matched pair $k=(a,b)$ in neighbor $j$'s matched set $P_j$, the implementation assigns the initial pair weight based on the matching cost $c_{ab}$
\begin{equation}
q^{(0)}_{j,k}
=
\exp\!\left(-\frac{c_{ab}}{T_q}\right),
\label{eq:app_q_init}
\end{equation}
which is used both for Kalman-prior initialization and as the pair weight entering pose refinement.

\textbf{IRLS pose solver.}
Both the initial pose estimation and the later pose-refinement rounds driven by Closed-Loop Posterior Scoring use weighted rigid alignment with IRLS. Following Eq.~(3) in the main text, at loop round $\ell$ the refined pose is estimated from the matched trajectory points by
\begin{equation}
\resizebox{0.98\linewidth}{!}{$\displaystyle
\Delta T_{j\to e}^{(\ell)}
=
\arg\min_{\Delta T\in SE(2)}
\sum_{k\in P_j}
q^{(\ell)}_{j,k}
\sum_{\kappa=0}^{L_p}
\rho_{\delta}\!\left(
\left\|
\Delta T\,p_{j,b}^{u_j-\kappa}
-
p_{e,a}^{u_j-\kappa}
\right\|_2
\right)
$},
\label{eq:app_irls}
\end{equation}
where each $k=(a,b)\in P_j$, $p_{j,b}^{u_j-\kappa}$ and $p_{e,a}^{u_j-\kappa}$ are the neighbor and ego box centers at the shared history offset $\kappa$, and $L_p=k_{\mathrm{hist}}$ in our implementation. Here $\rho_{\delta}$ is the Huber cost
\begin{equation}
\rho_{\delta}(r)=
\begin{cases}
\frac{1}{2}r^2, & r\le\delta,\\
\delta(r-\frac{1}{2}\delta), & r>\delta.
\end{cases}
\end{equation}
Thus, pair weight $q^{(\ell)}_{j,k}$ serves as the per-pair weight in IRLS, while $\rho_{\delta}$ provides robustness to outlier point residuals. In the released rule-based configuration, the solver uses 10 iterations, $\delta=0.5$, and convergence tolerance $10^{-4}$.

\textbf{Anchor state, prior covariance, and propagation.}
For a matched pair $k=(a,b)\in P_j$, the delayed anchor state is represented as
\begin{equation}
x^{u_j}_{j,k}=[p_x,p_y,v_x,v_y,\psi]^T.
\end{equation}
The planar velocity components $v_x$ and $v_y$ are expressed in meters per frame, obtained by converting physical velocities using the frame interval. Let $\tau_j$ denote the communication delay of neighbor $j$, so the current time satisfies $t=\tau_j+u_j$, and let $\Delta t_{\mathrm{frame}}$ denote the frame interval. For both datasets, $\Delta t_{\mathrm{frame}}=100$\,ms, corresponding to a sampling rate of 10\,Hz. We convert the delay to frame units as $\Delta n_j=\tau_j/\Delta t_{\mathrm{frame}}$. The initial covariance converted from $q^{(0)}_{j,k}$ is
\begin{equation}
\Sigma^{u_j}_{j,k}
=
\mathrm{diag}
\left(
\frac{\alpha_{p}}{q^{(0)}_{j,k}+\varepsilon_q},
\frac{\alpha_{p}}{q^{(0)}_{j,k}+\varepsilon_q},
\frac{\alpha_{v}}{q^{(0)}_{j,k}+\varepsilon_q},
\frac{\alpha_{v}}{q^{(0)}_{j,k}+\varepsilon_q},
\frac{(\alpha_{\psi}^{\mathrm{rad}})^2}{q^{(0)}_{j,k}+\varepsilon_q}
\right),
\label{eq:app_sigma_u}
\end{equation}
where $\alpha_{\psi}^{\mathrm{rad}}=(\pi/180)\alpha_{\psi}^{\mathrm{deg}}$ is the yaw scale expressed in radians.

The constant-velocity state transition used in the implementation is
\begin{equation}
A(\Delta n_j)=
\begin{bmatrix}
1 & 0 & \Delta n_j & 0 & 0\\
0 & 1 & 0 & \Delta n_j & 0\\
0 & 0 & 1 & 0 & 0\\
0 & 0 & 0 & 1 & 0\\
0 & 0 & 0 & 0 & 1
\end{bmatrix},
\end{equation}
and the process noise is
\begin{equation}
\begin{aligned}
Q(\Delta n_j)
&=
\mathrm{diag}\!\Big(
\sigma_a^2\max(\Delta n_j^2,1),
\sigma_a^2\max(\Delta n_j^2,1),\\
&\qquad
\sigma_a^2\max(\Delta n_j,1),
\sigma_a^2\max(\Delta n_j,1),\\
&\qquad
(\sigma_{\psi,Q}^{\mathrm{rad}})^2\max(\Delta n_j,1)
\Big).
\end{aligned}
\label{eq:app_q}
\end{equation}
The propagated prior state and covariance are then
\begin{equation}
x^-_{j,k}=A(\Delta n_j)x^{u_j}_{j,k},\qquad
\Sigma^-_{j,k}=A(\Delta n_j)\Sigma^{u_j}_{j,k}A(\Delta n_j)^T+Q(\Delta n_j).
\end{equation}

\begin{table*}[t]
    \centering
    \small
    \setlength{\tabcolsep}{3.5pt}
    \caption{Shared structural coefficients used in all experiments.}
    \label{tab:app_shared_rule_params}
    \renewcommand{\arraystretch}{1.06}
    \begin{tabular}{L{1.9cm}|L{2.8cm}|L{1.5cm}|L{7.3cm}}
        \toprule
        \textbf{Module} & \textbf{Parameter} & \textbf{Value} & \textbf{Role} \\
        \midrule
        Tracking & $k_{\mathrm{hist}}$ & 3 & Number of historical frames used to build short tracklets. \\
        \midrule
        \multirow{4}{*}{Matching} & $d_{\mathrm{gate}}$ & 4.0 & Position gate applied before a candidate pair is considered. \\
         & $\lambda_{\mathrm{vel}},\,\lambda_{\mathrm{graph}}$ & 1.0, 0.5 & Relative weights of the velocity and local-graph terms in Eq.~\eqref{eq:app_match_cost}. \\
         & $T_{\mathrm{vel}},\,T_{\mathrm{graph}}$ & 2.0, 3.0 & Normalizers for the velocity and graph terms in Eq.~\eqref{eq:app_match_cost}. \\
         & $K,\,w_{\mathrm{edge}},\,w_{\mathrm{pair}}$ & 3, 1.0, 0.5 & Neighbor count and edge/pair signature weights for local layout consistency. \\
        \midrule
        \multirow{2}{*}{Pose} & $N_{\mathrm{trk}}^{\min}$ & 3 & Minimum number of matched trajectories required before running pose refinement. \\
         & $n_{\mathrm{IRLS}},\,\delta,\,\epsilon_{\mathrm{IRLS}}$ & 10, 0.5, $10^{-4}$ & Shared Huber-IRLS configuration using pair weight $q^{(\ell)}_{j,k}$ as the per-pair weight. \\
        \midrule
        Statistical gate & $\tau_{\chi}$ & 7.815 & Fixed $\chi^2_{0.95}(3)$ gate for the 3-DoF observation $(x,y,\psi)$. \\
        \midrule
        Feedback residual & $|\mathcal{K}_{j,k}|$ & 3 & Number of historical frames used in the short-history term of Eq.~\eqref{eq:app_rhist}. \\
        \midrule
        \multirow{2}{*}{\makecell[l]{\footnotesize Closed-Loop\\\footnotesize Posterior Scoring}} & $N_{\mathrm{loop}}$ & 1 & Number of outer refinement rounds driven by Eq.~\eqref{eq:app_pose_feedback}. \\
         & $q_{\min}$ & $10^{-4}$ & Minimum pair weight retained for the next pose-refinement round after Eq.~\eqref{eq:app_pose_feedback}. \\
        \midrule
        Numerical constants & $\varepsilon_q$ & $10^{-6}$ & Small stabilizers in Eqs.~\eqref{eq:app_sigma_u}. \\
        \bottomrule
    \end{tabular}
\end{table*}

\begin{table*}[t]
    \centering
    \small
    \setlength{\tabcolsep}{3.5pt}
    \caption{Calibrated Kalman-style uncertainty parameters.}
    \label{tab:app_kalman_shared}
    \renewcommand{\arraystretch}{1.06}
    \begin{tabular}{L{2.40cm}|L{2.2cm}|L{1.45cm}|L{7.75cm}}
        \toprule
        \textbf{Group} & \textbf{Parameter} & \textbf{Value} & \textbf{Role} \\
        \midrule
        \multirow{4}{*}{Prior mapping} & $T_q$ & 1.5 & Cost-to-initial-weight temperature in Eq.~\eqref{eq:app_q_init}.\\
         & $\alpha_p$ & 0.1 & Position prior scale in Eq.~\eqref{eq:app_sigma_u}. \\
         & $\alpha_v$ & 0.03 & Velocity prior variance in Eq.~\eqref{eq:app_sigma_u}. \\
         & $\alpha_{\psi}^{\mathrm{deg}}$ & 3.0 & Yaw prior std factor in Eq.~\eqref{eq:app_sigma_u}. \\
        \midrule
        \multirow{2}{*}{Process noise $Q$} & $\sigma_a$ & 0.03 & Translational process noise in Eq.~\eqref{eq:app_q}. \\
         & $\sigma_{\psi,Q}^{\mathrm{deg}}$ & 4.0 & Yaw process noise in Eq.~\eqref{eq:app_q}. \\
        \midrule
        \multirow{2}{*}{\makecell[l]{Measurement\\noise $R$}} & $\sigma_{z,p}$ & 0.7 & Position observation noise in Eq.~\eqref{eq:app_hr}. \\
         & $\sigma_{z,\psi}^{\mathrm{deg}}$ & 8.0 & Yaw observation noise in Eq.~\eqref{eq:app_hr}. \\
        \midrule
        Feedback-decay coefficient & $\lambda_{\mathrm{fb}}$ & 1.5 & Feedback-decay coefficient in Eq.~\eqref{eq:app_pose_feedback}. \\
        \bottomrule
    \end{tabular}
\end{table*}

\textbf{Observation model, NIS gate, and update.}
The current-time ego observation measures only $(x,y,\psi)$, so the implementation uses
\begin{equation}
H=
\begin{bmatrix}
1&0&0&0&0\\
0&1&0&0&0\\
0&0&0&0&1
\end{bmatrix},
\qquad
R=
\mathrm{diag}
\left(
\sigma_{z,p}^2,\sigma_{z,p}^2,(\sigma_{z,\psi}^{\mathrm{rad}})^2
\right).
\label{eq:app_hr}
\end{equation}
Let $z^t_{j,k}=[z^t_{j,k,x},z^t_{j,k,y},z^t_{j,k,\psi}]^T$ be the current-time ego observation associated with pair $k$. Following Eq.~(6) in the main text, the measurement residual and its covariance are
\begin{equation}
\nu^t_{j,k}
=
\begin{bmatrix}
z^t_{j,k,x}-x^-_{j,k,x}\\
z^t_{j,k,y}-x^-_{j,k,y}\\
\mathrm{wrap}_{1/2}(z^t_{j,k,\psi}-x^-_{j,k,\psi})
\end{bmatrix},
\qquad
S^t_{j,k}=H\Sigma^-_{j,k}H^T+R.
\end{equation}
$\mathrm{wrap}_{1/2}(\cdot)$ denotes half-period yaw wrapping, i.e., mapping an angle difference to its principal value in the interval $[-\pi/2,\pi/2)$ under the $\pi$-periodic box-orientation ambiguity. This design choice is made to accommodate the common front-rear direction ambiguity in upstream 3D bounding box predictions. The normalized innovation squared is
\begin{equation}
d^{2}_{j,k}
=
\left(\nu^t_{j,k}\right)^T
\left(S^t_{j,k}\right)^{-1}
\nu^t_{j,k}.
\label{eq:app_d2}
\end{equation}
The measurement update is accepted only when $d^{2}_{j,k}\le\tau_{\chi}$, where we fix $\tau_{\chi}=\chi^2_{0.95}(3)=7.815$.

\textbf{Feedback residual and Closed-Loop Posterior Scoring.}
For each matched pair, the implementation first computes a short-history disagreement term over the delayed frame and several nearby history frames:
\begin{equation}
\begin{aligned}
r^{\mathrm{hist}}_{j,k}
\,&=
\frac{1}{|\mathcal{K}_{j,k}|}
\sum_{\kappa\in\mathcal{K}_{j,k}}
\Bigl(
\frac{\|\Delta T_{j\to e}^{(\ell)}p_{j,b}^{u_j-\kappa}-p_{e,a}^{u_j-\kappa}\|_2^2}{\sigma_{z,p}^2}
\\
&\qquad+
\frac{\mathrm{wrap}_{1/2}(\hat\psi_{j,b}^{u_j-\kappa,(\ell)}-\psi_{e,a}^{u_j-\kappa})^2}{(\sigma_{z,\psi}^{\mathrm{rad}})^2}
\Bigr),
\end{aligned}
\label{eq:app_rhist}
\end{equation}
where $\mathcal{K}_{j,k}$ contains the available history offsets and the delayed frame itself when $\Delta n_j>0$, and $\hat\psi_{j,b}^{u_j-\kappa,(\ell)}$ denotes the yaw of the neighbor box after applying the current loop-round transform $\Delta T_{j\to e}^{(\ell)}$. The feedback residual is then defined as
\begin{equation}
r^{\mathrm{fb}}_{j,k}=d^{2}_{j,k}+r^{\mathrm{hist}}_{j,k}.
\end{equation}
In Closed-Loop Posterior Scoring, the next-round pair weight is updated as
\begin{equation}
q^{(\ell+1)}_{j,k}
=
q^{(0)}_{j,k}\exp\!\left(-\lambda_{\mathrm{fb}}\,r^{\mathrm{fb}}_{j,k}\right).
\label{eq:app_pose_feedback}
\end{equation}
Pairs whose pair weight falls below the minimum threshold $q_{\min}$ are discarded before the next pose-refinement round.
The updated pair weight $q^{(\ell+1)}_{j,k}$ is then fed back to the IRLS pose solver in Eq.~\eqref{eq:app_irls} for the next round of spatial refinement, explicitly closing the spatio-temporal alignment loop.

\subsection{Shared and Calibrated Rule-based Parameters}

Table~\ref{tab:app_shared_rule_params} lists the structural coefficients that are fixed throughout all experiments. These parameters control the matching cost, local graph signatures, gating rule, and IRLS behavior. We keep them unchanged because their role is mainly structural rather than dataset-specific.

Table~\ref{tab:app_kalman_shared} lists the small set of Kalman-style uncertainty parameters used by the Kalman-style update, including the prior mapping, process noise, measurement noise, and closed-loop decay terms. Unlike the matching coefficients above, these parameters directly determine the uncertainty scale of delayed anchors and current-time observations, and are therefore the only groups that require light calibration.

\subsection{Parameter Calibration}

We calibrate the Kalman-style uncertainty parameters in Table~\ref{tab:app_kalman_shared} only once on the training split of V2V4Real, and then keep the resulting values unchanged in all reported experiments, including OPV2V and all test conditions. We choose V2V4Real as the calibration benchmark because it is the noisier and more realistic dataset, making it a more suitable reference point for setting the uncertainty scale of the object-level module.

The calibration is carried out under the representative \textbf{Joint-Hard} setting used in the main text, namely a communication delay of 200\,ms together with pose noise (0.6\,m, 0.6$^\circ$). This operating point reflects the coupled spatio-temporal perturbation regime that our method is primarily designed to handle, while avoiding benchmark-specific tuning across multiple conditions.

The calibration itself is intentionally lightweight. We first fix the statistical gate to $\chi^2_{0.95}(3)=7.815$, corresponding to a 95\% confidence gate for the 3-DoF observation $(x,y,\psi)$. We also use a target normalized innovation scale of 3 as the coarse initialization reference, since the current-time ego observation contains exactly three observed degrees of freedom. Starting from this reference scale, we then perform a small local sweep over the uncertainty-related groups in Table~\ref{tab:app_kalman_shared} on the V2V4Real training split, while keeping all matching and IRLS coefficients fixed.

The resulting parameter set is then reused unchanged on OPV2V and under all evaluation settings. Although this one-time calibration is not guaranteed to be strictly optimal for OPV2V, we observe no material degradation in our experiments. This suggests that the rule-based object-level module is not highly sensitive to benchmark-specific retuning once the uncertainty scale is set to a reasonable range.

Finally, the object-level closed-loop module itself contains no learnable parameters. The learned part of the overall system remains the detector and downstream fusion network described in the main text.

\subsection{Hyperparameter Sensitivity}

To verify that the reported gain is not tied to a narrow rule-based parameter choice, we vary the main structural and uncertainty-related hyperparameters individually on V2V4Real under \emph{Joint-Hard}. All other settings are held at their default values. As summarized in Table~\ref{tab:app-sensitivity}, all tested changes except the stricter 2\,m distance gate remain within 0.29 AP@0.5 of the default configuration. The result is consistent with the one-time calibration strategy above and indicates that the object-level module is insensitive to moderate parameter changes.

\begin{table}[!htbp]
    \centering
    \small
    \setlength{\tabcolsep}{4pt}
    \caption{Hyperparameter sensitivity on V2V4Real under \emph{Joint-Hard}. The default AP@0.5 is 67.04.}
    \label{tab:app-sensitivity}
    \begin{tabular}{lccc}
        \toprule
        \textbf{Parameter} & \textbf{Default} & \textbf{Test values} & \textbf{$\Delta$AP@0.5} \\
        \midrule
        Distance gate (m) & 4 & 2 / 6 & $-0.62$ / $-0.05$ \\
        History length (frames) & 3 & 2 / 4 & $-0.10$ / $+0.07$ \\
        NIS gate (\%) & 95 & 90 / 99 & $-0.08$ / $-0.07$ \\
        Feedback decay & 1.5 & 1.0 / 2.0 & $-0.08$ / $-0.04$ \\
        Huber penalty & 0.5 & 0.25 / 1.0 & $-0.17$ / $-0.29$ \\
        Minimum solver support & 3 & 4 / 5 & $-0.10$ / $-0.15$ \\
        \bottomrule
    \end{tabular}
\end{table}

\FloatBarrier

\section{Additional Qualitative Results}
\label{sec:app-qualitative}

\begin{figure}[!htbp]
    \centering
    \begin{minipage}[t]{\linewidth}
        \centering
        \includegraphics[width=\linewidth]{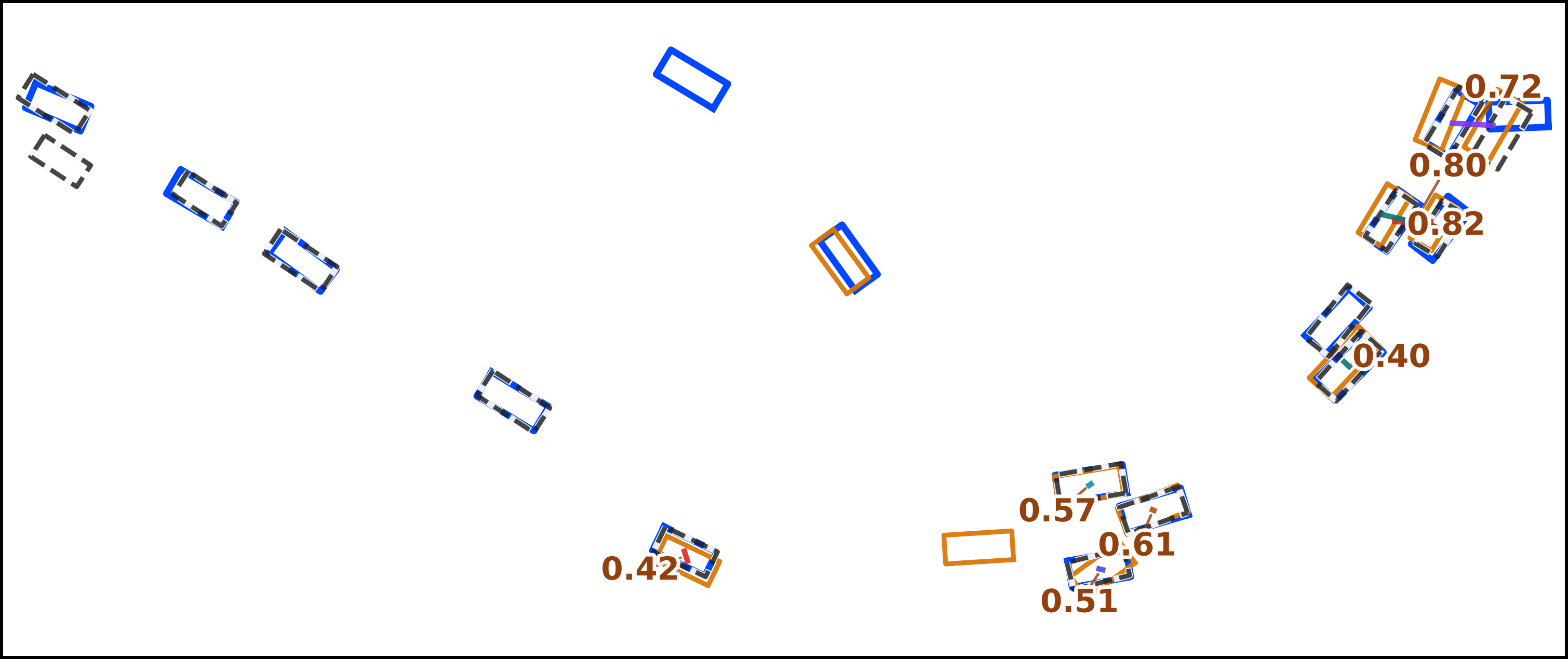}\par
        \vspace{-0.8mm}
        {\footnotesize (a) After initial pose refinement (at delayed time $u_j=t-\tau_j$)\par}
        \vspace{0.9mm}
    \end{minipage}

    \begin{minipage}[t]{\linewidth}
        \centering
        \includegraphics[width=\linewidth]{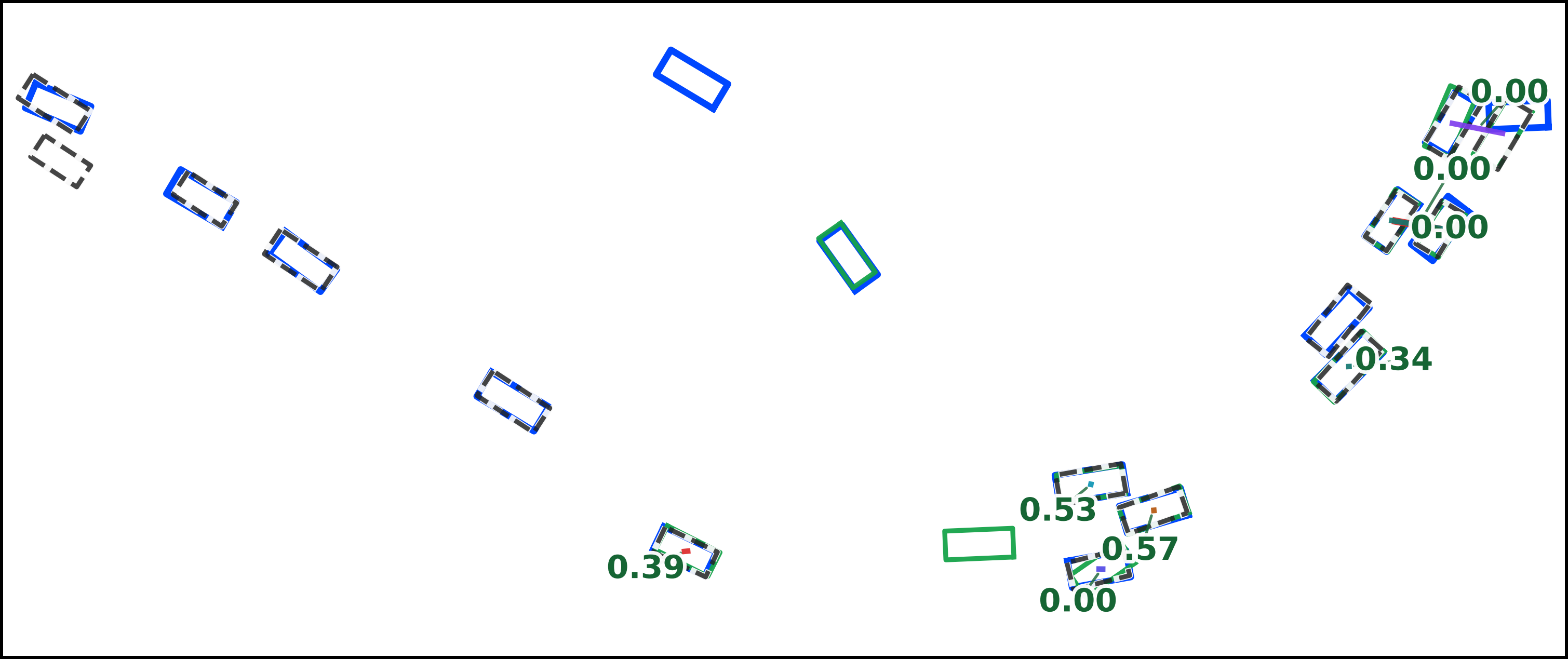}\par
        \vspace{-0.8mm}
        {\footnotesize (b) After Closed-Loop Posterior Scoring (at delayed time $u_j=t-\tau_j$)\par}
        \vspace{0.9mm}
    \end{minipage}

    \begin{minipage}[t]{\linewidth}
        \centering
        \includegraphics[width=\linewidth]{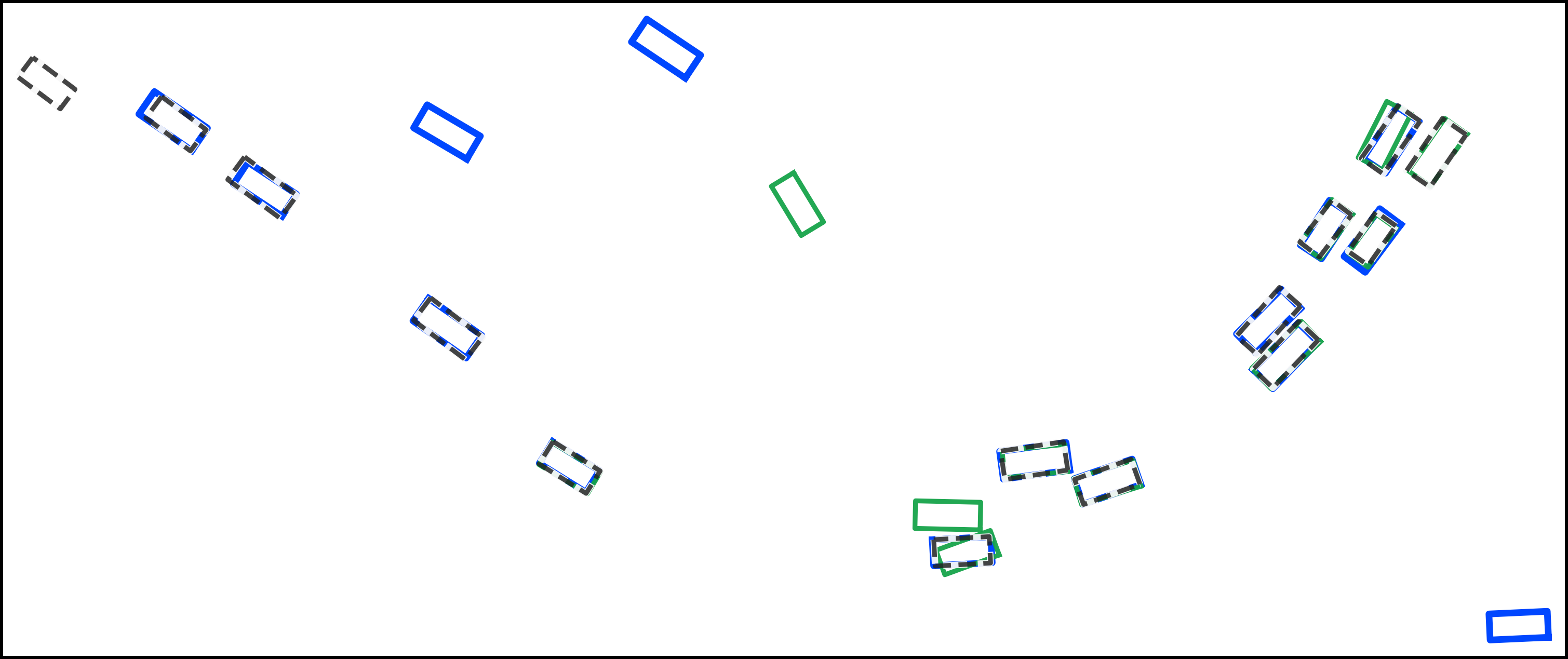}\par
        \vspace{-0.8mm}
        {\footnotesize (c) After current-time propagation (at current time $t$ after closed-loop)\par}
    \end{minipage}
    \vspace{0.8mm}

    {\scriptsize
        \setlength{\tabcolsep}{0pt}
        \renewcommand{\arraystretch}{1.05}
        \begin{tabular*}{0.98\linewidth}{@{\extracolsep{\fill}}p{0.48\linewidth}p{0.48\linewidth}@{}}
            \makebox[2.2em][l]{\raisebox{-0.7ex}{\tikz[baseline=-0.6ex]{
                \draw[orange!90!black,line width=0.95pt] (0,0) rectangle (0.34,0.18);
            }}} Initial-refined neighbor box &
            \makebox[2.2em][l]{\raisebox{-0.7ex}{\tikz[baseline=-0.6ex]{
                \draw[green!60!black,line width=0.95pt] (0,0) rectangle (0.34,0.18);
            }}} Closed-loop-refined neighbor box \\
        \end{tabular*}\par\vspace{-0.8ex}
        \begin{tabular*}{0.98\linewidth}{@{\extracolsep{\fill}}p{0.31\linewidth}p{0.31\linewidth}p{0.31\linewidth}@{}}
            \makebox[2.2em][l]{\raisebox{-0.7ex}{\tikz[baseline=-0.6ex]{
                \draw[blue!70!black,line width=0.95pt] (0,0) rectangle (0.34,0.18);
            }}} Ego box &
            \makebox[2.2em][l]{\raisebox{-0.7ex}{\tikz[baseline=-0.6ex]{
                \draw[black,dashed,line width=0.95pt] (0,0) rectangle (0.34,0.18);
            }}} Ground-truth box &
            \makebox[2.2em][l]{\raisebox{-0.15ex}{\textsf{0.87}}} pair weight
        \end{tabular*}}
    \caption{Representative qualitative example of the object-level pipeline in \name. This example visualizes how \name progressively improves delayed collaborator alignment through object-level pose refinement, Closed-Loop Posterior Scoring, and temporal propagation.}
    \label{fig:app_failure_pose_refinement}
    \Description{Three vertically arranged bird's-eye-view panels show the same OPV2V scene after initial pose refinement, after closed-loop suppression of unreliable matched pairs, and after propagation to the current ego time. Neighbor boxes become better aligned with ego and ground-truth boxes across the three stages, although some residual motion error remains.}
\end{figure}

Figure~\ref{fig:app_failure_pose_refinement} visualizes three representative stages of our object-level pipeline.
All three panels are taken from an OPV2V example under a 400\,ms communication delay and pose noise $(0.6\,\text{m}, 0.6^\circ)$.

Together, the panels illustrate both the closed-loop reliability update and the remaining limitation of the simple rule-based propagation model.
The panel (a) shows the initial pose-refinement result before Closed-Loop Posterior Scoring.
In this example, the estimated pose is still poor, and several mismatched pairs remain because the delayed collaborator boxes are affected jointly by inaccurate detections and initial relative-pose error.
This case illustrates the limitation of relying only on one-shot delayed-frame matching and pose refinement: when the input correspondences are already corrupted, the resulting alignment can still be unreliable even if the optimization itself is well defined.

The panel (b) shows the pose after Closed-Loop Posterior Scoring.
Compared with the initial result, the refined alignment becomes much cleaner and most corresponding boxes are brought into close agreement.
Importantly, four previously matched pairs are assigned extremely small pair weight values (below $10^{-2}$), which effectively suppresses their influence in the next pose-refinement round.
This example directly reflects the role of Closed-Loop Posterior Scoring: instead of trusting all delayed-frame correspondences equally, the method re-evaluates them using current-time agreement and short-history consistency, and then downweights pairs that are no longer trustworthy.

The panel (c) shows the propagated collaborator boxes after rewriting and temporal propagation to the current ego time.
Most delayed neighbor boxes are now well aligned with the ego-side observations, indicating that the object-level propagation is effective in bringing stale collaborator evidence closer to the current frame.
At the same time, several boxes still exhibit residual offsets rather than perfect overlap.
This limitation is expected: under delays of several hundred milliseconds, assuming approximately constant-velocity straight-line motion is a reasonable and computationally efficient approximation, but it cannot fully model all real object motions.
As a result, while the propagation stage is practically useful and aligns most objects well, its simplicity can still become a bottleneck in some hard cases.

\FloatBarrier

\section{Extended Experimental Results}
\label{sec:app-extended-results}

\begin{figure*}[t]
    \centering
    \begin{minipage}[t]{0.24\textwidth}
        \centering
        \includegraphics[width=\linewidth]{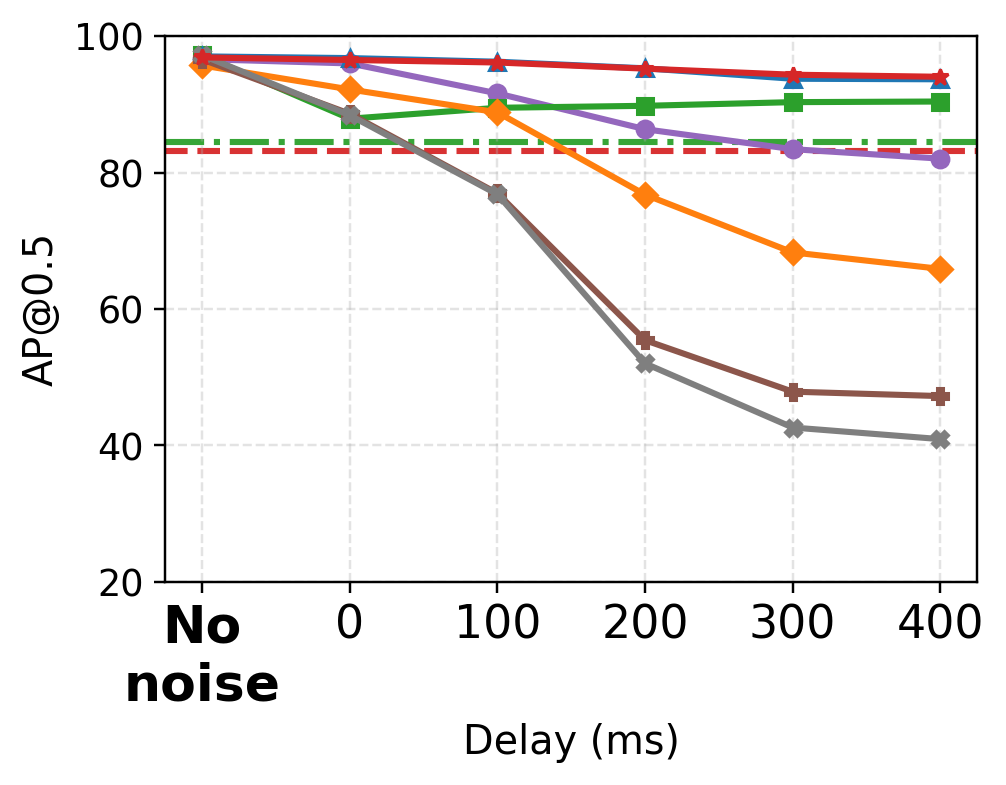}
        \vspace{-1mm}
        {\hspace*{0.13\linewidth}\parbox[t]{0.87\linewidth}{\centering\small (a) OPV2V, fixed pose std = 0.3}}
    \end{minipage}\hfill
    \begin{minipage}[t]{0.24\textwidth}
        \centering
        \includegraphics[width=\linewidth]{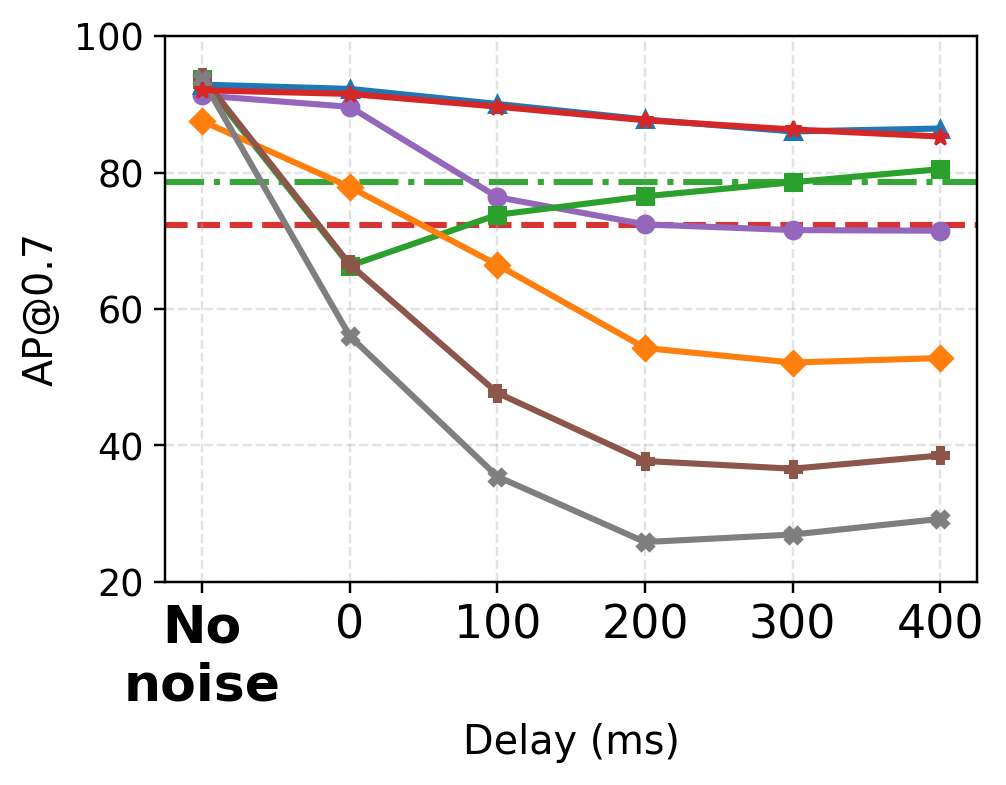}
        \vspace{-1mm}
        {\hspace*{0.13\linewidth}\parbox[t]{0.87\linewidth}{\centering\small (b) OPV2V, fixed pose std = 0.3}}
    \end{minipage}\hfill
    \begin{minipage}[t]{0.24\textwidth}
        \centering
        \includegraphics[width=\linewidth]{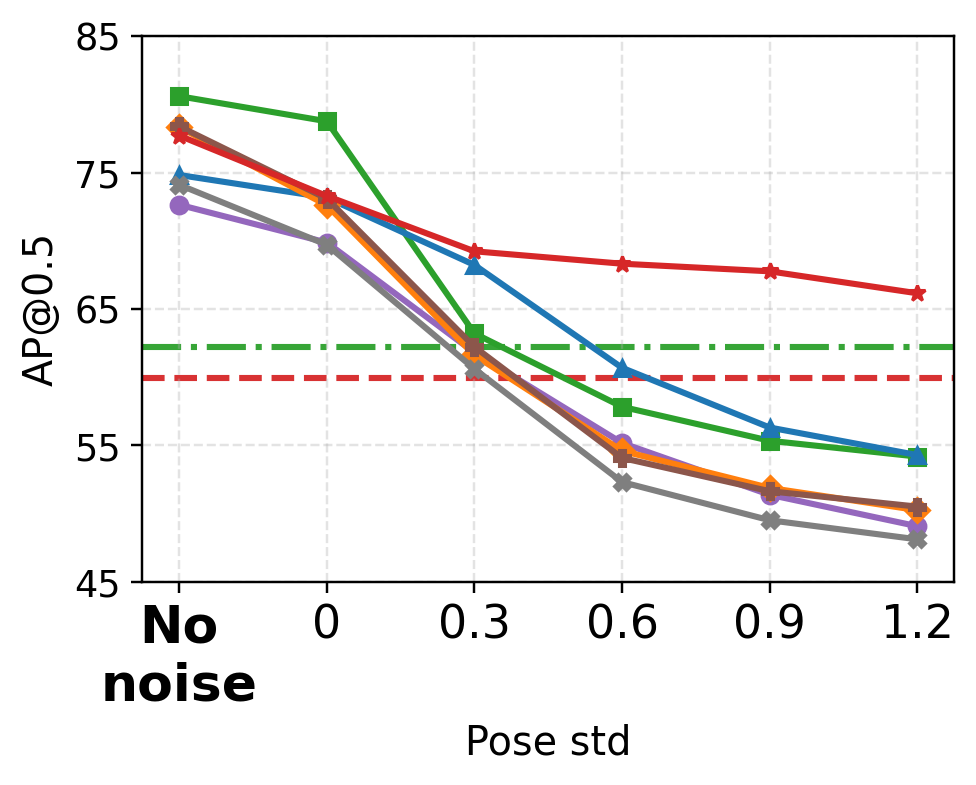}
        \vspace{-1mm}
        {\parbox[t]{\linewidth}{\centering\fontsize{8.4pt}{9.2pt}\selectfont (c) V2V4Real, fixed delay = 100ms}}
    \end{minipage}\hfill
    \begin{minipage}[t]{0.24\textwidth}
        \centering
        \includegraphics[width=\linewidth]{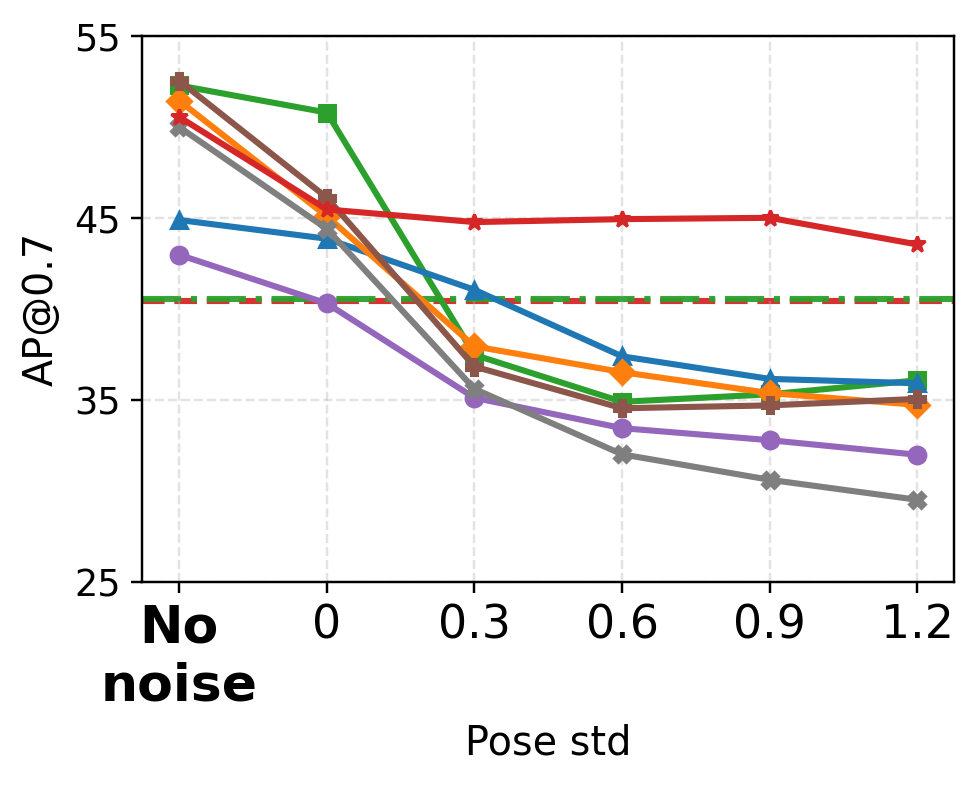}
        \vspace{-1mm}
        {\parbox[t]{\linewidth}{\centering\fontsize{8.4pt}{9.2pt}\selectfont (d) V2V4Real, fixed delay = 100ms}}
    \end{minipage}

    \vspace{1mm}
    {\small
        \setlength{\tabcolsep}{0pt}
        \renewcommand{\arraystretch}{1.05}
        \hspace*{0.01\textwidth}\begin{tabular*}{0.98\textwidth}{@{\extracolsep{\fill}}>{\raggedright\arraybackslash}p{0.18\textwidth}>{\raggedright\arraybackslash}p{0.18\textwidth}>{\raggedright\arraybackslash}p{0.18\textwidth}>{\raggedright\arraybackslash}p{0.18\textwidth}>{\raggedright\arraybackslash}p{0.18\textwidth}@{}}
            \makebox[2.9em][l]{\raisebox{-0.7ex}{\tikz[baseline=-0.6ex]{
                        \draw[red!80!black,line width=1.1pt] (0,0.09) -- (0.34,0.09);
                        \draw[red!80!black,line width=1.1pt] (0.17,0.03) -- (0.17,0.15);
                        \draw[red!80!black,line width=1.1pt] (0.11,0.09) -- (0.23,0.09);
                        \draw[red!80!black,line width=1.1pt] (0.125,0.045) -- (0.215,0.135);
                        \draw[red!80!black,line width=1.1pt] (0.125,0.135) -- (0.215,0.045);
            }}} \name &
            \makebox[2.9em][l]{\raisebox{-0.7ex}{\tikz[baseline=-0.6ex]{
                        \draw[green!60!black,line width=1.1pt] (0,0.09) -- (0.34,0.09);
                        \fill[green!60!black] (0.13,0.05) rectangle (0.21,0.13);
            }}} TraF-Align &
            \makebox[2.9em][l]{\raisebox{-0.7ex}{\tikz[baseline=-0.6ex]{
                        \draw[blue!70!black,line width=1.1pt] (0,0.09) -- (0.34,0.09);
                        \fill[blue!70!black] (0.17,0.15) -- (0.11,0.03) -- (0.23,0.03) -- cycle;
            }}} \base &
            \makebox[2.9em][l]{\raisebox{-0.7ex}{\tikz[baseline=-0.6ex]{
                        \draw[orange!90!black,line width=1.1pt] (0,0.09) -- (0.34,0.09);
                        \fill[orange!90!black] (0.17,0.15) -- (0.23,0.09) -- (0.17,0.03) -- (0.11,0.09) -- cycle;
            }}} V2X-ViT &
            \\
        \end{tabular*}\par\vspace{-1.2ex}
        \noindent\hspace*{0.01\textwidth}\begin{tabular*}{0.98\textwidth}{@{\extracolsep{\fill}}>{\raggedright\arraybackslash}p{0.18\textwidth}>{\raggedright\arraybackslash}p{0.18\textwidth}>{\raggedright\arraybackslash}p{0.18\textwidth}>{\raggedright\arraybackslash}p{0.18\textwidth}>{\raggedright\arraybackslash}p{0.18\textwidth}@{}}
            \makebox[2.9em][l]{\raisebox{-0.7ex}{\tikz[baseline=-0.6ex]{
                        \draw[brown!70!black,line width=1.1pt] (0,0.09) -- (0.34,0.09);
                        \draw[brown!70!black,line width=1.1pt] (0.17,0.03) -- (0.17,0.15);
                        \draw[brown!70!black,line width=1.1pt] (0.11,0.09) -- (0.23,0.09);
            }}} ERMVP &
            \makebox[2.9em][l]{\raisebox{-0.7ex}{\tikz[baseline=-0.6ex]{
                        \draw[gray!70!black,line width=1.1pt] (0,0.09) -- (0.34,0.09);
                        \draw[gray!70!black,line width=1.1pt] (0.12,0.04) -- (0.22,0.14);
                        \draw[gray!70!black,line width=1.1pt] (0.12,0.14) -- (0.22,0.04);
            }}} CoST &
            \makebox[2.9em][l]{\raisebox{-0.7ex}{\tikz[baseline=-0.6ex]{
                        \draw[draw={rgb,255:red,148;green,103;blue,189},line width=1.1pt] (0,0.09) -- (0.34,0.09);
                        \fill[fill={rgb,255:red,148;green,103;blue,189}] (0.17,0.09) circle (0.04);
            }}} CoAlign &
            \makebox[2.9em][l]{\raisebox{-0.7ex}{\tikz[baseline=-0.6ex]{
                        \draw[red!80!black,dashed,line width=1.1pt] (0,0.09) -- (0.34,0.09);
            }}} \name ego-only &
            \makebox[2.9em][l]{\raisebox{-0.7ex}{\tikz[baseline=-0.6ex]{
                        \draw[green!60!black,dash dot,line width=1.1pt] (0,0.09) -- (0.34,0.09);
            }}} TraF ego-only
            \\
        \end{tabular*}}
    \caption{Additional robustness curves omitted from the main text. Comparison with state-of-the-art methods on OPV2V and V2V4Real. Each entry is reported as AP@0.5 / AP@0.7.}
    \label{fig:app_additional_robustness_curves}
    \Description{Four line charts complement the robustness curves in the main text. The first two show OPV2V average precision as communication delay increases under fixed pose noise. The last two show V2V4Real average precision as pose noise increases under a fixed communication delay.}
\end{figure*}

\subsection{Additional Motivation Results}

Table~\ref{tab:app_moti_result} provides the detailed measurement results for the pilot study discussed in Sec.~\ref{sec_moti} of the main text.
\begin{table}[H]
    \centering
    \small
    \setlength{\tabcolsep}{4.2pt}
    \caption{Additional motivation measurements. Method comparison under fixed 100ms communication delay with varying Gaussian pose noise.}
    \label{tab:app_moti_result}
    \vspace{-2mm}
    \resizebox{\columnwidth}{!}{
    \begin{tabular}{l|ccc}
        \toprule
        \multicolumn{4}{c}{\textbf{AP@0.5 / AP@0.7 in OPV2V} (100ms fixed)} \\
        \midrule
        \textbf{Method} & \textbf{None} & \textbf{$(0.3\,\text{m}, 0.3^\circ)$} & \textbf{$(0.6\,\text{m}, 0.6^\circ)$} \\
        \hline
        CoAlign \cite{coalign} & 95.24 / 80.50 & 91.61 / 76.39 & 88.52 / 74.90 \\
        TraF-Align (ego-only) & 84.49 / 78.57 & 84.49 / 78.57 & 84.49 / 78.57 \\
        TraF-Align \cite{traf_align} & \textbf{96.72 / 92.12} & 89.50 / 73.81 & 84.10 / 68.95 \\
        \base \cite{coalign,traf_align} & 96.58 / 91.55 & \textbf{96.26 / 90.06} & \textbf{89.89 / 80.55} \\
        \bottomrule
    \end{tabular}}

    \vspace{1mm}

    \resizebox{\columnwidth}{!}{
    \begin{tabular}{l|ccc}
        \toprule
        \multicolumn{4}{c}{\textbf{AP@0.5 / AP@0.7 in V2V4Real} (100ms fixed)} \\
        \midrule
        \textbf{Method} & \textbf{None} & \textbf{$(0.3\,\text{m}, 0.3^\circ)$} & \textbf{$(0.6\,\text{m}, 0.6^\circ)$} \\
        \hline
        CoAlign \cite{coalign} & 69.86 / 40.30 & 61.60 / 34.93 & 55.16 / 33.44 \\
        TraF-Align (ego-only) & 62.19 / 40.55 & 62.19 / 40.55 & \textbf{62.19 / 40.55} \\
        TraF-Align \cite{traf_align} & \textbf{78.76 / 50.80} & 63.20 / 37.47 & 57.82 / 34.89 \\
        \base \cite{coalign,traf_align} & 73.16 / 43.86 & \textbf{68.22 / 41.05} & 60.68 / 37.39 \\
        \bottomrule
    \end{tabular}}
\end{table}

The OPV2V results indicate that the cascaded baseline \base can be beneficial on the cleaner simulated benchmark, although it does not yield a gain in every case: at zero pose noise, it is slightly below TraF-Align (96.58 / 91.55 vs.\ 96.72 / 92.12), but once the pose perturbation increases to 0.3 and 0.6, it becomes clearly stronger than TraF-Align (96.26 / 90.06 vs.\ 89.50 / 73.81, and 89.89 / 80.55 vs.\ 84.10 / 68.95).
This suggests that under relatively clean detections and correspondences, explicit spatial refinement can still provide a beneficial upstream reference for temporal compensation, but its benefit is conditional rather than unconditionally guaranteed.

On V2V4Real, the pattern is consistent with the measurement study in the main text: direct cascading does not provide a stable gain over TraF-Align, and under stronger pose perturbation it can even fall below the ego-only reference. This further supports that simply stacking spatial pose refinement and temporal compensation is insufficient in realistic scenes.

\subsection{Additional Robustness Curves}

Figure~\ref{fig:app_additional_robustness_curves} complements the robustness analysis in the main text by adding the four omitted sweep directions. Together with the corresponding figure in the main text, they complete the delay/noise sweeps on both datasets.

Here we focus on one specific phenomenon: the non-monotonic delay behavior of TraF-Align.
A mild version of this pattern is already visible in the V2V4Real delay sweep in the main text, and the OPV2V delay sweep in Fig.~\ref{fig:app_additional_robustness_curves}(a)--(b) makes it much clearer.
Under fixed pose noise, TraF-Align first drops substantially as delay increases from the no-noise point, but then partially rebounds when the delay becomes even larger, especially at AP@0.7.

One possible explanation is that TraF-Align explicitly encodes delay in the collaborator branch, so the model can learn to reduce the effective weight of neighbor information once the delay becomes very large.
This suppression may reduce false positives introduced by stale collaborator features, and the effect is particularly visible at AP@0.7, where small localization biases are more likely to turn a seemingly plausible prediction into a mismatch.
However, this strategy is still implicit and coarse.
It is not tied to an explicit object-level notion of trustworthiness, and it reduces collaborator influence in a relatively non-selective manner.
As a result, it may suppress some harmful stale evidence, but it also discards many useful collaborative cues and therefore weakens the potential collaborative gain.

The curves also suggest that under coupled pose and delay perturbations, pose noise cannot be handled merely by reducing the influence of collaborator information.
In the V2V4Real pose-noise sweep of Fig.~\ref{fig:app_additional_robustness_curves}(c)--(d), our method maintains comparatively robust performance as the pose noise increases under fixed delay.
A reason is that our method does not treat collaborator information in an all-or-nothing way.
Instead, it propagates delayed hypotheses at the object level and re-checks them against current-time ego observations, so that collaborator evidence can be filtered more selectively according to its current spatial trustworthiness.

\subsection{High-Motion Subset Analysis}

We further examine the constant-velocity prior on a high-motion subset of V2V4Real under the \emph{Joint-Hard} setting, which uses a 200\,ms communication delay and pose noise $(0.6\,\mathrm{m},0.6^\circ)$. The subset is selected exclusively from \emph{ground-truth object trajectories}, rather than predictions from any evaluated method, so all methods are evaluated on exactly the same frames. A frame is selected if any co-visible ground-truth object has a yaw change larger than $15^\circ$ across adjacent annotated frames, or a center residual larger than $0.5$\,m when its two preceding ground-truth states are used to form a constant-velocity prediction of the current state. This procedure selects 379 out of 1,993 test frames (19.0\%).

\begin{table}[H]
    \centering
    \small
    \setlength{\tabcolsep}{5pt}
    \caption{Results on the V2V4Real high-motion subset under \emph{Joint-Hard}. Each entry is AP@0.5 / AP@0.7.}
    \label{tab:app-high-motion}
    \begin{tabular}{l|cc}
        \toprule
        \textbf{Method} & \textbf{All frames} & \textbf{High-motion subset} \\
        \midrule
        TraF-Align~\cite{traf_align} & 58.18 / 35.09 & 56.35 / 33.50 \\
        \base~\cite{coalign,traf_align} & 61.04 / 37.69 & 58.83 / 36.01 \\
        \textbf{\name} & \textbf{67.04 / 44.05} & \textbf{62.76 / 40.63} \\
        \bottomrule
    \end{tabular}
\end{table}

As shown in Table~\ref{tab:app-high-motion}, the selected subset is harder for all methods. Nevertheless, \name retains a clear advantage: it outperforms TraF-Align by 6.41 / 7.13 AP and \base by 3.93 / 4.62 AP on the high-motion subset. The result indicates that nonlinear motion degrades the constant-velocity prior but does not eliminate the benefit of object-level verification: unreliable innovations can still be rejected by the statistical gate or assigned small closed-loop weights before they dominate pose refinement.

\end{document}